\documentclass{article}

\usepackage{arxiv}

\usepackage[utf8]{inputenc} % allow utf-8 input
\usepackage[T1]{fontenc}    % use 8-bit T1 fonts
\usepackage{hyperref}       % hyperlinks
\usepackage{url}            % simple URL typesetting
\usepackage{booktabs}       % professional-quality tables
\usepackage{amsfonts}       % blackboard math symbols
\usepackage{nicefrac}       % compact symbols for 1/2, etc.
\usepackage{microtype}      % microtypography
\usepackage{lipsum}
\usepackage{graphicx}
\graphicspath{ {./images/} }
\usepackage{natbib} 
\usepackage[table]{xcolor}
\usepackage{subfigure}
\usepackage{subcaption}
\usepackage{enumitem}  
\usepackage{fancyvrb}

\newcommand{\SemAP}{$\text{SemAP}$}
\newcommand{\SemAPObs}{$\text{SemAP}_{\text{Phe}}$}
\newcommand{\SemAPRea}{$\text{SemAP}_{\text{Rea}}$}
\newcommand{\SemAPFull}{$\text{SemAP}_{\text{Full}}$}
\newcommand{\SemFOne}{$\text{SemF1}$}
\newcommand{\SemFOneObs}{$\text{SemF1}_{\text{Phe}}$}
\newcommand{\SemFOneRea}{$\text{SemF1}_{\text{Rea}}$}
\newcommand{\SemFOneFull}{$\text{SemF1}_{\text{Full}}$}

\newcommand{\CSemAPObs}{$\text{CSemAP}_{\text{Phe}}$}
\newcommand{\CSemAPRea}{$\text{CSemAP}_{\text{Rea}}$}
\newcommand{\CSemAPFull}{$\text{CSemAP}_{\text{Full}}$}

\newcommand{\CSemFOneObs}{$\text{CSemF1}_{\text{Phe}}$}
\newcommand{\CSemFOneRea}{$\text{CSemF1}_{\text{Rea}}$}
\newcommand{\CSemFOneFull}{$\text{CSemF1}_{\text{Full}}$}

\usepackage{amsmath,epstopdf,amssymb,color,url,multirow,multicol,verbatim,threeparttable,diagbox,makecell,booktabs,colortbl,graphicx}
\usepackage{algorithm,pifont}
\usepackage{algpseudocode}

\definecolor{light-gray}{gray}{0.82}
\definecolor{lightblue}{RGB}{100, 149, 237}
\definecolor{lightgreen}{RGB}{60, 179, 113}

\title{Semantic Visual Anomaly Detection and Reasoning in AI-Generated Images}

\author{
Chuangchuang Tan$^{1,2}$\thanks{Work done during internship at MSRA.}\quad 
Xiang Ming$^{2}$\quad
Jinglu Wang$^{2}$\quad 
Renshuai Tao$^{1}$\quad
Bin Li$^{3}$\quad \\
\textbf{Yunchao Wei}$^{1}$ 
\; \textbf{Yao Zhao$^{1}$}\quad
\textbf{Yan Lu$^{2}$}\quad\\[1.2mm]
$^1$Beijing Jiaotong University \quad
$^2$Microsoft Research Asia \quad
$^3$Shenzhen University 
}

\begin{document}
\maketitle

\begin{abstract}

The rapid advancement of 
    AI-generated content (AIGC) has enabled the synthesis of visually convincing images; however, many such outputs exhibit subtle \textbf{semantic anomalies}, including unrealistic object configurations, violations of physical laws, or commonsense inconsistencies, which compromise the overall plausibility of the generated scenes.
Detecting these semantic-level anomalies 
    is essential for assessing the trustworthiness of AIGC media, especially in AIGC image analysis, explainable deepfake detection and semantic authenticity assessment.
In this paper, 
    we formalize \textbf{semantic anomaly detection and reasoning} for AIGC images and 
    introduce \textbf{AnomReason}, a large-scale benchmark with structured annotations as quadruples \emph{(Name, Phenomenon, Reasoning, Severity)}. 
Annotations are produced by 
    a modular multi-agent pipeline (\textbf{AnomAgent}) with lightweight human-in-the-loop verification, enabling scale while preserving quality.
    At construction time, AnomAgent processed approximately 4.17\,B GPT-4o tokens, providing scale evidence for the resulting structured annotations.
We further 
    show that models fine-tuned on AnomReason achieve consistent gains over strong vision-language baselines under our proposed semantic matching metric (\textit{SemAP} and \textit{SemF1}).
    Applications to {explainable deepfake detection} and {semantic reasonableness assessment of image generators} demonstrate practical utility.
In summary, AnomReason and AnomAgent 
    serve as a foundation for measuring and improving the semantic plausibility of AI-generated images.
We will release code, metrics, data, and task-aligned models to support reproducible research on semantic authenticity and interpretable AIGC forensics.

\end{abstract}

\section{Introduction}

% 段落1：背景与动机，强调AIGC语义异常问题及其对关键应用的影响，吸引关注
The rapid advancement of 
    AI-generated content (AIGC) has led to striking progress in photorealistic image synthesis, powered by large-scale generative models such as Stable Diffusion~\citep{rombach2022high}, DALL·E~\citep{ramesh2021zero}, Midjourney~\citep{midjourney2025}, and Flux~\citep{flux2024}. 
These models can generate 
    high-quality images and are being widely adopted in design, education, media, and science. 
However, 
    despite visual realism, many AIGC-generated images exhibit subtle but significant \emph{\textbf{semantic-level anomalies}}, such as \textbf{logical contradictions}, \textbf{physical implausibilities}, or \textbf{commonsense violations}, that compromise their authenticity. 
These inconsistencies highlight 
    the need for structured semantic anomaly detection in AIGC images, 
    which not only facilitates a deeper analysis of such images but also lays the groundwork for explainable deepfake detection and evaluating generative models' reasoning capabilities regarding commonsense knowledge. Furthermore, this approach holds potential for enhancing the semantic coherence of image generation models.
Our proposed task directly 
    supports these goals by identifying content-aware violations that compromise semantic authenticity.

\begin{figure}[ht!]
\begin{center}
    \includegraphics[width=\linewidth]{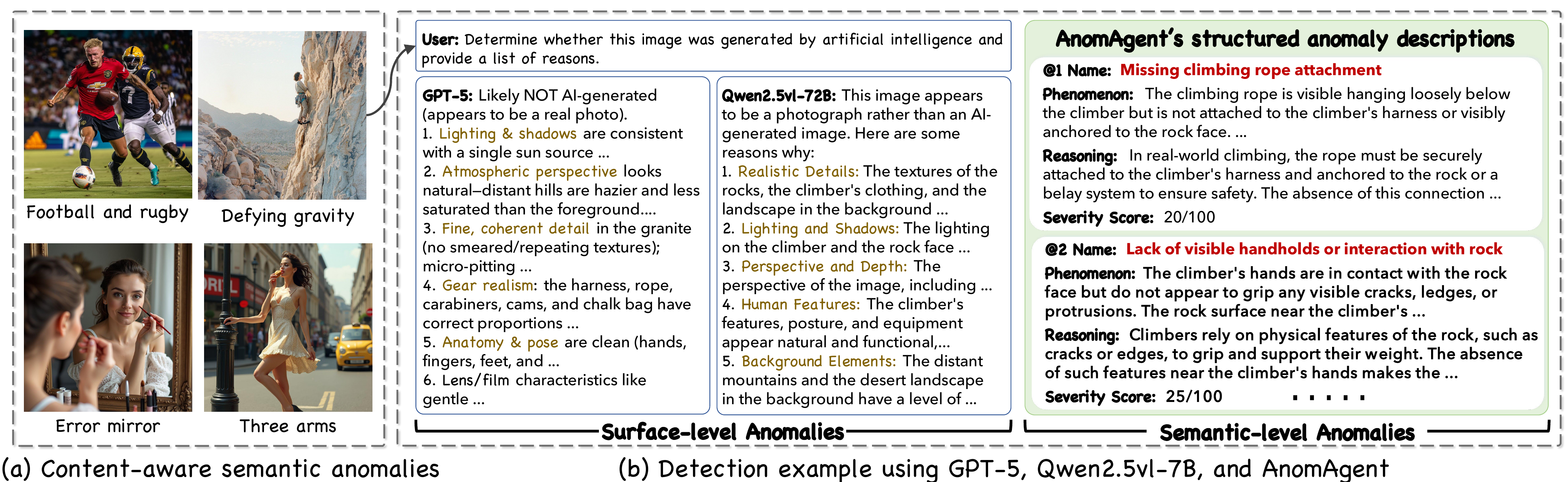}
\end{center}
\caption{
\textbf{Semantic anomaly detection in AIGC-generated images.}
\textbf{(a)} Illustration of high-level semantic anomalies that are context-dependent and subtle, such as inconsistent physics, anatomy, and reflections—challenges that go beyond surface-level visual artifacts.
\textbf{(b)} Comparison of detection performance between general-purpose vision-language models (e.g., GPT-5, Qwen2.5vl-72B) and the proposed \textbf{AnomAgent}. While the former focus on surface-level cues such as lighting and textures, AnomAgent identifies fine-grained semantic inconsistencies and provides structured, explainable outputs with severity ratings.
}
\label{fig:aigc_abnormal_examples}
\end{figure}

% 段落2：配图强化语义异常的多样性和难以检测性，引出挑战
As illustrated in Fig.~\ref{fig:aigc_abnormal_examples}(a), 
    such semantic anomalies include hybrid semantics (e.g., mixing sports equipment),
    violations of physics (e.g., gravity-defying climbers), 
    and anatomical implausibilities. 
These issues 
    are not captured by traditional low-level forensic cues~\citep{Tan2023CVPR,tan2024rethinking, durall2020watch, wang2020cnn, ojha2023towards}, yet they critically affect human trust and decision-making when AIGC is used in factual or sensitive domains.

This work introduces 
    the task of \textbf{semantic visual anomaly detection and reasoning} for AIGC image, 
    which seeks to identify and explain \emph{semantic-level anomalies} present in synthetic images. 
These anomalies pertain specifically 
    to violations of commonsense knowledge, physical plausibility, and logical coherence.
Formally defined, 
    given an AIGC-generated image as input, 
    the system is required to produce a set of structured anomaly descriptions comprising four components: 
    \textbf{\textit{Name}}, 
    \textbf{\textit{Phenomenon}}, 
    \textbf{\textit{Reasoning}}, 
    and a corresponding \textbf{\textit{Severity Score}}, capturing \emph{what} is wrong, \emph{why} it is wrong, and \emph{how severe} it is, as illustrated in Fig.~\ref{fig:aigc_abnormal_examples}(b). 
    Specifically, \textit{Name} provides a concise summary of the anomaly, while \textit{Phenomenon} offers a detailed description at the semantic level. \textit{Reasoning} explains the underlying causes of the anomaly. Finally, \textit{Severity Score} quantifies the anomaly by assigning a score that reflects its authenticity.
This formulation 
    underscores not only the importance of detecting anomalies 
    but also providing explanations alongside detailed semantic evaluations.

% 这么改的话就不是与现有 task 对比了
Several studies ~\citep{wen2025spot,gao2025fakereasoning,zhang2025ivy,zhou2025aigi} 
    have attempted to extract forgery-related evidence using vision-language models (VLMs)~\citep{bai2025qwen2} from images to support classification outcomes.
However, 
    their performance on anomaly detection frequently relies on surface-level irregularities, 
    such as global lighting conditions or shadow patterns—subtle statistical artifacts in texture that are typically imperceptible to human observers 
    (see Fig.~\ref{fig:aigc_abnormal_examples}(b), Left).
In contrast, 
    our task focuses on \emph{content-level semantic anomalies}, implausible object interactions, physical violations, or commonsense errors, that are directly visible to humans and therefore more aligned with human judgment (refer to Fig.~\ref{fig:aigc_abnormal_examples}(b), Right).
Furthermore, instead of shallow descriptions, the defined \emph{structured anomaly representation} makes anomaly analysis interpretable and accessible, moving beyond detection to structured reasoning.

% 段落6：引出Benchmark AnomReason，作为任务的核心资源支撑
To support this task, 
    we build \textbf{AnomReason}, 
    the first large-scale benchmark for content-aware semantic anomaly detection in AIGC images. 
AnomReason consists of 
    diverse synthetic scenes annotated with structured semantic anomalies. 
Each anomaly entry describes what is wrong, 
    why it is wrong, and how severe the inconsistency is—capturing errors in object composition, spatial arrangement, interaction logic, or physical constraints. 
This benchmark is enabled by \textbf{AnomAgent}, 
    a modular multi-agent framework that decomposes anomaly reasoning into object perception, attribute analysis, relational reasoning, and anomaly synthesis. 
To ensure the reliability of annotations produced by AnomAgent, 
    we incorporate a lightweight \textbf{human-in-the-loop verification} stage. 
This hybrid pipeline 
    balances scale and accuracy by filtering and refining automatically generated results. 
Compared with purely manual annotation or fully automated generation, 
    our multi-agent plus human verification strategy achieves both interpretability and scalability, 
    allowing AnomReason to provide high-quality structured annotations at unprecedented scale.
Furthermore, 
    we propose novel anomaly semantic detection metrics based on Average Precision (AP) and F1-score, 
    referred to as SemAP and SemF1, 
    respectively, to facilitate the evaluation of anomaly semantic detection performance.

By shifting the focus from surface-level artifacts to content-level reasoning, 
    our framework opens new research directions in semantic authenticity, explainable forensics, and commonsense reasoning for generative media. 
Our proposed benchmark and system, 
    AnomReason and AnomAgent, enable several critical applications:
    (i) training semantic anomaly detector to gain deeper understanding AIGC images, 
    (ii) developing \textit{explainable deepfake detectors} that not only classify but also provide semantic-level justifications for their predictions, 
    and (iii) conducting \textit{AIGC semantic reasonableness assessment} to evaluate the logical coherence of image generation models. 
These applications strengthen content authenticity auditing and help guide future AIGC model development.

\section{AnomAgent Framework}
\label{sec:anomagent}

Detecting semantic anomalies 
    in AI-generated images requires not only visual recognition 
    but also reasoning about commonsense knowledge, physical feasibility, and multi-object interactions. 
To address this challenge, 
    we propose \textbf{AnomAgent}, a modular multi-agent framework designed to emulate human perception and reasoning through structured processes.

\begin{figure}[t]
    \centering
    \includegraphics[width=\linewidth]{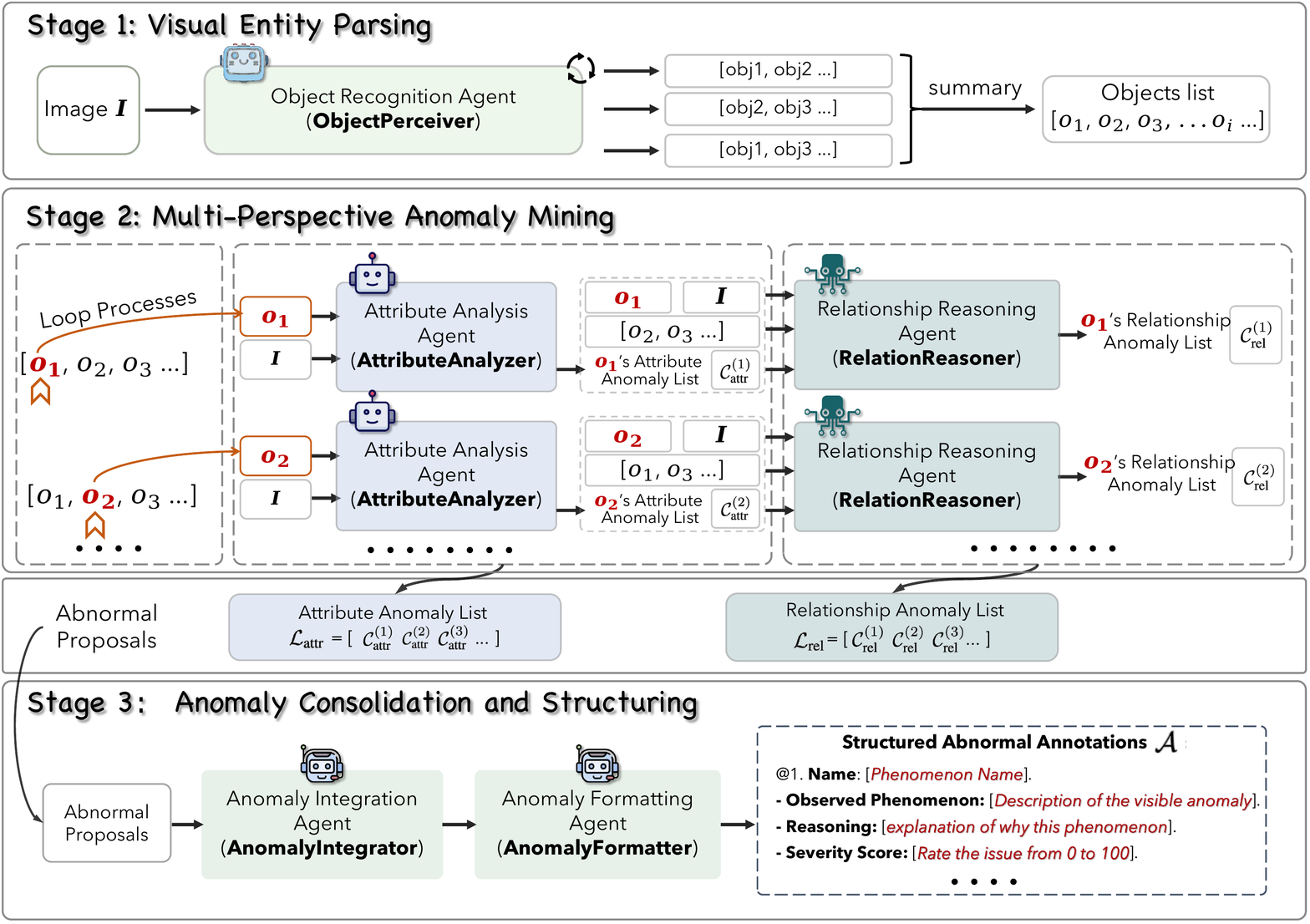}
    \caption{
    \textbf{Overview of the \textit{AnomAgent} pipeline for semantic anomaly annotation.}
    {Stage~1} parses visual entities and yields an object list $\mathcal{O}$.
    {Stage~2} performs multi-perspective anomaly mining, producing attribute candidates $\mathcal{C}_{\text{attr}}$ and relational candidates $\mathcal{C}_{\text{rel}}$, which are scored and pruned to $\mathcal{C}^{+}$.
    {Stage~3} consolidates candidates (merging near-duplicates to $\hat{\mathcal{C}}$) and outputs structured anomalies $\mathcal{A}=\{(y,o,r,v)\}$ (Name, Phenomenon, Reasoning, Severity).
    }
    \label{fig:anomagent_pipeline}
\end{figure}

As shown in Fig.~\ref{fig:anomagent_pipeline}, 
    AnomAgent decomposes the anomaly detection process into three stages: entity parsing, anomaly mining, and structured output generation. 
Each stage involves 
    specialized agents that collaborate to produce interpretable, high-precision semantic anomaly annotations.

Given an input image $I$, AnomAgent outputs a set of structured anomalies:
\begin{equation}
\mathcal{A} = \left\{ (y_i, o_i, r_i, v_i) \right\}_{i=1}^m
\end{equation}
where $y_i$ is 
    the anomaly name, 
    $o_i$ the described anomaly phenomenon,
    $r_i$ the reasoning explains why $o$ is considered anomalous, 
    and $v_i \in [0, 100]$ indicates the severity.    
A score of 
    $0$ denotes implausibility, while $100$ represents full realism.

\subsection{Stage 1: Visual Entity Parsing}

Semantic inconsistencies 
    are often object-centric. However, objects in AIGC images can be entangled, distorted, or hallucinated—making entity extraction unreliable.
The \textbf{Object Recognition Agent} (\texttt{ObjectPerceiver}) identifies all semantically distinct entities in the image, with emphasis on human-related objects.

To reduce false negatives, the detection is repeated $T$ times with varying prompts and merged:
\begin{equation}
\mathcal{O}^{(t)} = \texttt{ObjectPerceiver}(I), \quad t = 1, \dots, T; \quad \mathcal{O} = \bigcup_{t=1}^{T} \mathcal{O}^{(t)}
\end{equation}
Each object $o_i \in \mathcal{O}$ is represented by an object name and a detailed description.

\subsection{Stage 2: Multi-Perspective Anomaly Mining}
Anomalies in AIGC images 
    may arise from incorrect attributes or implausible inter-object interactions. 
In this stage, 
    we iterate over each object $o_i \in \mathcal{O}$ and perform two complementary forms of semantic consistency analysis: intra-object (attribute-based) and inter-object (relation-based).

\textbf{Attribute-Level Analysis.}
The \textbf{Attribute Analysis Agent} (\texttt{AttributeAnalyzer}) 
    examines visual attributes of $o_i$ such as shape, material, and functionality. 
It identifies internal inconsistencies and produces a set of attribute anomaly candidates:
\begin{equation}
\mathcal{C}_{\text{attr}}^{(i)} = \texttt{AttributeAnalyzer}(o_i)
\end{equation}

\textbf{Relationship-Level Analysis.}  
The \textbf{Relationship Reasoning Agent} (\texttt{RelationReasoner}) 
    evaluates spatial, semantic, and functional interactions between $o_i$ and the rest of the scene, guided by its own attribute anomalies $\mathcal{C}_{\text{attr}}^{(i)}$ as contextual priors:
\begin{equation}
\mathcal{C}_{\text{rel}}^{(i)} = \texttt{RelationReasoner}\left(o_i, \mathcal{O} \setminus \{o_i\}, \mathcal{C}_{\text{attr}}^{(i)}\right)
\end{equation}
The agent 
    first enumerates pairwise and groupwise relations, then filters semantically implausible ones using both visual context and object-specific inconsistencies.
We define the intermediate anomaly lists across all objects as:
$
\mathcal{L}_{\text{attr}} = \left\{ \mathcal{C}_{\text{attr}}^{(i)} \right\}_{i=1}^{|\mathcal{O}|}, \quad
\mathcal{L}_{\text{rel}} = \left\{ \mathcal{C}_{\text{rel}}^{(i)} \right\}_{i=1}^{|\mathcal{O}|}
$
The total set of candidate anomalies is:
$
\mathcal{C} = \bigcup_{i=1}^{|\mathcal{O}|} \left( \mathcal{C}_{\text{attr}}^{(i)} \cup \mathcal{C}_{\text{rel}}^{(i)} \right).
$
In addition, 
    to mitigate hallucinations and reduce abnormal omissions, we implement a two-step process in AttributeAnalyzer and RelationReasoner. 
First, anomalies are comprehensively identified from multiple perspectives; 
subsequently, these anomalies are verified and structurally outputted.

\subsection{Stage 3: Anomaly Consolidation and Structuring}
Raw anomaly candidates are noisy, redundant, and linguistically inconsistent. Annotations require clean, standardized outputs.
We next consolidate and structure the raw candidate set $\mathcal{C}$ into interpretable anomaly annotations.

\textbf{Integration.}  
The \textbf{Anomaly Integration Agent} (\texttt{AnomalyIntegrator}) consolidates overlapping or redundant candidates and removes noise:
$
\hat{\mathcal{C}} = \texttt{AnomalyIntegrator}(\mathcal{C}).
$

\begin{figure}[t]
    \centering
    \includegraphics[width=\linewidth]{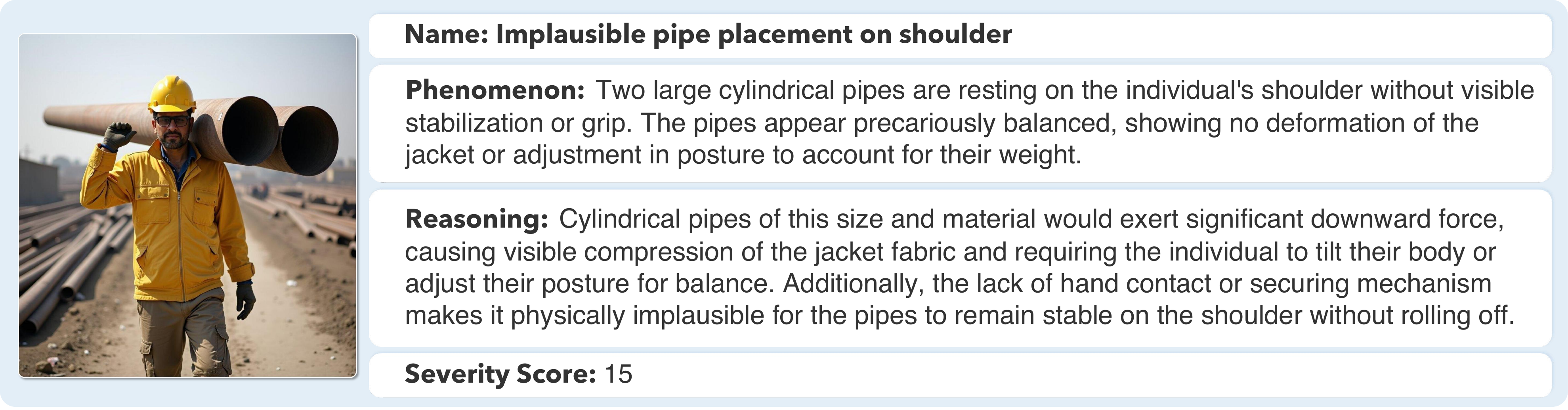}
    \caption{ 
    \textbf{Example of structured anomalies.} This figure illustrates a detected anomaly where two cylindrical pipes are unrealistically balanced on the individual’s shoulder. By structuring the anomaly as \{Name, Observed Phenomenon, Reasoning, Severity Score\}, the model not only provides a clear description of the anomaly but also offers an interpretable reasoning process, making it easier to understand why this arrangement is physically implausible. The severity score quantifies the degree of implausibility, enhancing the model's ability to observe and explain semantic-level anomalies. This structure allows for transparent and interpretable anomaly detection, improving the detection model's trustworthiness and explainability.
    }
    \label{fig:structured_candidate_anomalies}
\end{figure}

\textbf{Formatting.}  
The \textbf{Anomaly Formatting Agent} (\texttt{AnomalyFormatter}) maps each anomaly candidate $c \in \hat{\mathcal{C}}$ into a structured four-field annotation:
$
\mathcal{A} = \left\{ (y_i, o_i, r_i, v_i) \right\}_{i=1}^{|\hat{\mathcal{C}}|}.
$
In Fig.~\ref{fig:structured_candidate_anomalies}, we present examples of structured anomalies.
This structured output enables applications such as quality assessment and deepfake detection.

AnomAgent provides 
    a modular, interpretable, and scalable solution to semantic anomaly detection. 
By decomposing 
    the reasoning process across multiple agents, it aligns with human judgment and supports practical applications such as quality auditing and explainable detection. 
Additional details about AnomAgent are provided in Appendix~\ref{sec:agent_details}.

\section{AnomReason Benchmark}
\label{sec:anomreason}

Despite growing interest in 
    semantic anomaly detection, there is no standardized benchmark designed to evaluate \emph{semantic-level} anomalies in AIGC content. 
We construct \textbf{AnomReason}, 
    a large-scale dataset of photorealistic AIGC images annotated with structured semantic anomalies across attribute, relational, and commonsense dimensions.

\subsection{Data Construction}
We collect a diverse set of 
    image-text pairs by crawling approximately 600K user prompts and their corresponding outputs from Midjourney~\citep{midjourney2025}. 
To ensure content realism and diversity, 
    we apply CLIP-based~\citep{radford2021learning} filtering on embedding alignment, 
    extracting 109,058 visually realistic samples. 
A subset of 9,911 images is 
    further manually verified for semantic richness and authenticity.

To enhance generative diversity, 
    we synthesize additional samples using Stable Diffusion~3.5~\citep{stabilityai2024sd35} and Flux~\citep{flux2024}, 
    using the same prompt pool. 
After automated and manual filtering, 
    we construct a photorealistic dataset comprising 21,539 images: 9,911 from Midjourney, 4,645 from SD3.5, and 6,983 from Flux.

\subsection{Automatic annotation with AnomAgent}
\label{subsec:auto-anno}
We apply the 
    GPT-4o-based \textbf{AnomAgent} framework (Sec.~\ref{sec:anomagent}) on the full image set, 
    producing 174,872 structured candidate anomalies across 21,539 images.
Across the full corpus, 
    the pipeline consumed about 4.17 billion GPT-4o tokens to generate and refine candidate anomalies before HITL screening.\footnote{We report tokens as aggregated API usage logs, including both prompt and completion.}
Each anomaly contains 
    a textual name, observed phenomenon, commonsense-based reasoning, and a severity rating in $[0,100]$.

\begin{figure}[t]
    \centering
    \includegraphics[width=\linewidth]{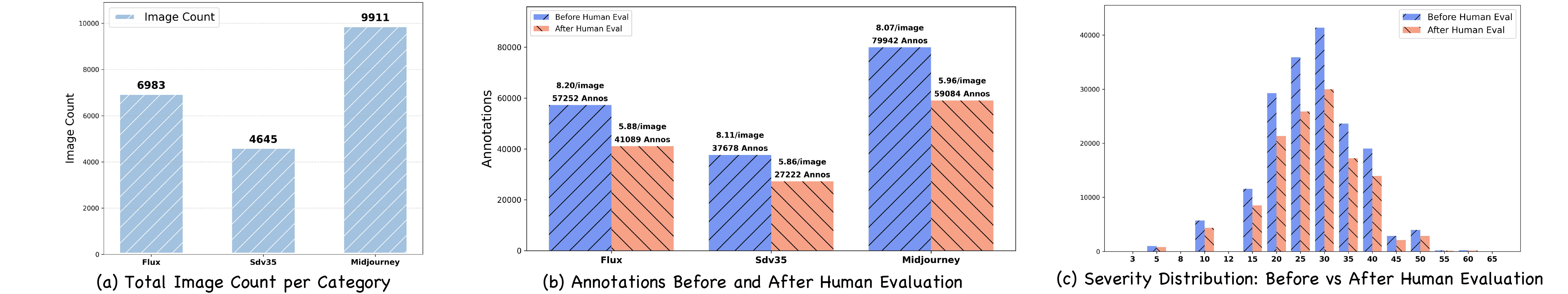}
    \caption{
    \textbf{AnomReason statistics.} (a) Total image count per category: Flux contains 6983 images, Sdv3.5 contains 4645, and Midjourney has the most with 9911 images. (b) Annotations before and after human evaluation: Flux has a reduction from 8.20 to 5.88 annotations per image, Sdv3.5 decreases from 8.11 to 5.86, and Midjourney shows a slight drop from 8.07 to 5.96 annotations per image. (c) Severity distribution before and after human evaluation: It showing a shift towards lower severity values after human evaluation, reflecting the refinement process in the annotation quality.
    }
    \label{fig:anomreason_stats}
\end{figure}

To ensure annotation reliability, 
    we introduce a lightweight \textit{human-in-the-loop (HITL)} quality control stage. 
Each candidate anomaly $a$ 
    is screened by a trained annotator answering a single-choice question:  
\textit{“Is this structured description correct for the given image?``}  
with three options: \textsc{accept}/\textsc{reject}/\textsc{unsure}.
\begin{equation}
h(a)\in\{1,0,\bot\}\quad\text{for }\textsc{accept},\ \textsc{reject},\ \textsc{unsure}, 
\qquad 
\mathcal{A}_{\text{final}} = \{\,a\in\mathcal{A}:\ h(a)=1\,\}.
\end{equation}
This low-cost protocol 
    removes implausible hallucinations while preserving structured quality. 
After HITL filtering, 
    the average valid annotations per image drops from ~8 to ~5.9, indicating refined semantic focus (Fig.~\ref{fig:anomreason_stats}).

\textbf{Dataset Split.}
We retain \textbf{21,533} images for training/testing after removing six duplicates/corrupted items.
We adopt 
    a deterministic 50/50 train/test split ($|\mathcal{D}| = 21{,}533$), 
    comprising 10,765 images for training and 10,774 for testing. 
Each split includes 
    image-level metadata, 
    anomaly counts, 
    severity scores, 
    and intermediate agent outputs to support replicability and ablation studies. 
We will release 
    all results encompassing intermediate outputs from AnomAgent nodes alongside structured annotations filtered through HIL processing mechanisms.

\subsection{Evaluation protocol}
\label{subsec:eval}

We propose 
    a structure-aware evaluation protocol tailored for semantic-level anomaly reasoning. 
Each annotation tuple $(y, o, r, v)$ 
    contains a description $o$ and a reasoning field $r$, 
    which are compared against ground truth using BERTScore~\citep{bert-score}. 
We define 
    three evaluation views: \textbf{Phe} (Phenomenon), \textbf{Rea} (Reasoning), and \textbf{Full} (combined fusion).

We perform 
    one-to-one anomaly matching at the image level based on similarity thresholds $\tau \in \{0.7, 0.8, 0.9\}$, and compute precision/recall curves to derive:
\begin{equation}
\text{SemAP}_v = \frac{1}{|\mathcal{D}|} \sum_{I \in \mathcal{D}} \text{AP}_v(I),\quad
\text{SemF1}_v = \frac{1}{|\mathcal{D}|} \sum_{I \in \mathcal{D}} \text{F1}_v(I)
\end{equation}

Severity scores $v$ 
    can optionally serve as confidence scores in ranking tasks. 
We adopt 
    the ``distilbert-base-uncased'' model for computing BertScore. 
Comprehensive details are provided in Appendix~\ref{appendix:sem-metrics}.

\section{Experiments}

We conduct comprehensive experiments 
    to evaluate the effectiveness of our benchmark, methodology, and task formulation in three progressively structured settings.
These experiments 
    are designed to assess both the capabilities of current VLMs and the benefits of targeted fine-tuning for semantic anomaly analysis.
First, 
    we evaluate structured semantic anomaly detection and reasoning on the \textit{AnomReason} benchmark, 
    measuring whether VLMs can both identify anomalous phenomena and explain their semantic violations (Sec.~\ref{subsec:main-results}). 
Second, 
    we extend this reasoning capability to a more applied setting: 
    explainable deepfake detection on \textit{AnomReason-Deepfake}, 
    which requires accurate AI-generated content classification and grounded explanation (Sec.~\ref{subsec:deepfake}). 
Third, 
    we audit modern text-to-image generators by quantifying semantic plausibility and anomaly severity in their outputs using structured reasoning (Sec.~\ref{subsec:reasonableness}).

\begin{table}[htp!]
\centering
\caption{Comparative performance on the \textit{AnomReason-Test}.}
\label{tab:AIGC_SVADR_results}
\resizebox{\textwidth}{!}{
\begin{tabular}{l ccc|ccc}
\toprule
\textbf{Model} & \textbf{\SemAPObs} & \textbf{\SemAPRea} & \textbf{\SemAPFull} & \textbf{\SemFOneObs} & \textbf{\SemFOneRea} & \textbf{\SemFOneFull} \\
\midrule
LongVA-7B~\citep{zhang2024longva}                           & 0.2579 & 0.2593 & 0.2494 & 0.1641 & 0.1617 & 0.1578 \\
LLaVA-OV-7B~\citep{li2024llava}                             & 0.3280 & 0.2987 & 0.3012 & 0.2044 & 0.1837 & 0.1867 \\
Phi-3.5-Vision~\citep{abdin2024phi3technicalreporthighly}   & 0.3466 & 0.3040 & 0.3117 & 0.2960 & 0.2572 & 0.2647 \\
MiniCPM-V-2.6~\citep{huminicpm}                             & 0.3898 & 0.3501 & 0.3537 & 0.1891 & 0.1698 & 0.1715 \\
InternVL2-8B~\citep{wang2024mpo}                            & 0.3697 & 0.3424 & 0.3424 & 0.3789 & 0.3500 & 0.3507 \\
InternVL2.5-8B~\citep{chen2024expanding}                    & 0.3343 & 0.3070 & 0.3087 & 0.3008 & 0.2733 & 0.2764 \\
InternVL3-8B~\citep{zhu2025internvl3}                       & 0.4552 & 0.3676 & 0.3927 & 0.1948 & 0.1614 & 0.1703 \\
InternVL3-9B~\citep{zhu2025internvl3}                       & 0.3871 & 0.3371 & 0.3514 & 0.3953 & 0.3456 & 0.3595 \\
mPLUG-Owl3-7B~\citep{ye2024mplug}                           & 0.4026 & 0.3661 & 0.3678 & 0.1247 & 0.1141 & 0.1144 \\
Qwen2-VL-7B ~\citep{wang2024qwen2}                          & 0.4090 & 0.3564 & 0.3678 & 0.1307 & 0.1208 & 0.1207 \\
InternVL2-26B~\citep{wang2024mpo}                           & 0.4209 & 0.3728 & 0.3865 & 0.4048 & 0.3590 & 0.3722 \\
Qwen2.5-VL-7B~\citep{bai2025qwen2}                          & 0.4674 & 0.3902 & 0.4155 & 0.4240 & 0.3548 & 0.3775 \\
Qwen2.5-VL-72B~\citep{bai2025qwen2}                         & 0.4926 & 0.4353 & 0.4568 & 0.4423 & 0.3912 & 0.4104 \\
\midrule
Gemini-2.5-pro~\citep{google_gemini25pro_preview_2025}      & 0.3755 & 0.3127 & 0.3384 & 0.1995 & 0.1674 & 0.1806 \\
GPT-o3~\citep{openai2024o3}                                 & 0.4470 & 0.3762 & 0.4058 & 0.4431 & 0.3724 & 0.4021 \\
GPT-5~\citep{openai_gpt5_system_card_2025}                  & 0.3760 & 0.3212 & 0.3469 & 0.4407 & 0.3762 & 0.4065 \\
GPT-4o~\citep{hurst2024gpt}                                 & 0.4908 & 0.4562 & 0.4727 & \textbf{0.5304} & 0.4930 & \textbf{0.5109} \\
\midrule
\rowcolor{gray!10}
AnomReasonor-7B                                             &\textbf{ 0.5221} & \textbf{0.5130} & \textbf{0.5162} & 0.5066 & \textbf{0.4977} & 0.5009 \\

\bottomrule
\end{tabular}
}
\end{table}

\subsection{AIGC Semantic Visual Anomaly Detection and Reasoning}
\label{subsec:main-results}
The first task evaluates
    whether VLMs can detect and explain semantic anomalies using structured outputs. 
The core challenge lies in 
    not only localizing errors but also providing plausible reasoning chains grounded in commonsense or physical priors.
We adopt the 
    AnomReason \texttt{test} set and use \textbf{SemAP} and \textbf{SemF1} metrics across phenomenon (Phe), reasoning (Rea), and full (Phe+Rea) views. We fine-tune Qwen2.5-VL-7B via LoRA on the \texttt{train} split, resulting in a baseline \textbf{AnomReasonor-7B (AR-7B)}. (training details in App.~\ref{app:training})

\textbf{Results.}
Table~\ref{tab:AIGC_SVADR_results} presents 
    comprehensive results on semantic anomaly detection and reasoning. 
Overall, most off-the-shelf VLMs 
    struggle with the full task (\SemAPFull), with values typically below 0.42, indicating limited semantic understanding in the absence of targeted supervision. 
Among open-source models, 
    \textbf{Qwen2.5-VL-72B} performs best (\SemAPFull = 0.4568), 
    followed by InternVL3-8B and InternVL2-26B. 
However, they still lag behind 
    the top-performing \textit{AnomReasonor-7B}, which achieves a new state-of-the-art \textbf{\SemAPFull = 0.5162} and the highest reasoning score \textbf{\SemAPRea = 0.5130},
    demonstrating the effectiveness of fine-tuning on our structured anomaly supervision. 
Notably, \textit{AnomReasonor-7B} also surpasses GPT-4o in all \textbf{\SemAP} metrics.
In terms of \textbf{\SemFOne} (alignment quality), 
    GPT-4o retains a slight edge (\SemFOneFull = 0.5109), but \textit{AnomReasonor-7B} is highly competitive (\SemFOneFull = 0.5009), particularly excelling in reasoning (\SemFOneRea = 0.4977 vs. 0.4930 for GPT-4o). 
This highlights 
    that our fine-tuned model closes the gap to proprietary systems not only in detection but also in the quality of generated descriptions.

Interestingly, 
    most models show stronger anomaly observation (\SemAPObs) than reasoning (\SemAPRea), 
    with some (e.g., InternVL3-8B) showing a wide gap (0.4552 vs. 0.3676), suggesting that identifying ``something wrong'' is easier than articulating ``why''. 
In contrast, \textit{AnomReasonor-7B} 
    exhibits the most balanced profile, with reasoning performance nearly matching observation. 
This reflects 
    the benefit of our structured annotation format and supervision signal, which jointly improve both detection and explanation.

\begin{table}[h]

\centering
\caption{Explainable deepfake detection on \textit{AnomReason-Deepfake}. 
}

\label{tab:deepfake_results}
\resizebox{\textwidth}{!}{
\begin{tabular}{l c|ccc|ccc}
\toprule
\textbf{Models} &  \textbf{Acc} & \textbf{\CSemAPObs} & \textbf{\CSemAPRea} & \textbf{\CSemAPFull} & \textbf{\CSemFOneObs} & \textbf{\CSemFOneRea} & \textbf{\CSemFOneFull} \\
\midrule
LongVA-7B~\citep{zhang2024longva}     & 53.45  & 0.0722 & 0.0714 & 0.0688 & 0.0458 & 0.0448 & 0.0436 \\
LLaVA-OV-7B~\citep{li2024llava}              & 66.26  & 0.1235 & 0.1124 & 0.1141 & 0.0818 & 0.0738 & 0.0752 \\
Phi-3.5-Vision~\citep{abdin2024phi3technicalreporthighly}        & 41.33  & 0.0685 & 0.0602 & 0.0616 & 0.0570 & 0.0497 & 0.0511 \\
MiniCPM-V-2.6~\citep{huminicpm}        & 55.05  & 0.0233 & 0.0205 & 0.0209 & 0.0109 & 0.0100 & 0.0100 \\
InternVL2-8B~\citep{wang2024mpo}         & 59.46  & 0.0739 & 0.0702 & 0.0697 & 0.0720 & 0.0680 & 0.0678 \\
InternVL2.5-8B~\citep{chen2024expanding}       & 63.61  & 0.1165 & 0.1085 & 0.1089 & 0.0988 & 0.0911 & 0.0919 \\
InternVL3-8B~\citep{zhu2025internvl3}         & 62.84  & 0.2949 & 0.2394 & 0.2559 & 0.1240 & 0.1031 & 0.1089 \\
InternVL3-9B~\citep{zhu2025internvl3}         & 64.04  & 0.2293 & 0.1987 & 0.2081 & 0.2343 & 0.2041 & 0.2132 \\
mPLUG-Owl3-7B~\citep{ye2024mplug}        & 55.96  & 0.0445 & 0.0411 & 0.0404 & 0.0130 & 0.0120 & 0.0117 \\
Qwen2-VL-7B~\citep{wang2024qwen2}          & 67.21  & 0.1710 & 0.1421 & 0.1496 & 0.0524 & 0.0458 & 0.0467 \\
InternVL2-26B~\citep{wang2024mpo}        & 54.73  & 0.0194 & 0.0172 & 0.0179 & 0.0180 & 0.0160 & 0.0166 \\
Qwen2.5-VL-7B~\citep{bai2025qwen2}        & 65.41  & 0.1295 & 0.1085 & 0.1155 & 0.1124 & 0.0943 & 0.1004 \\
Qwen2.5-VL-72B~\citep{bai2025qwen2}         & 77.60  & 0.2626 & 0.2337 & 0.2453 & 0.2427 & 0.2159 & 0.2266 \\
\midrule
Gemini-2.5-pro~\citep{google_gemini25pro_preview_2025}     & 85.65 & 0.2631 & 0.2192 & 0.2382 & 0.1488 & 0.1249 & 0.1351 \\
GPT-o3~\citep{openai2024o3}                & 85.60 & 0.3189 & 0.2690 & 0.2898 & 0.3187 & 0.2685 & 0.2895 \\
GPT-5~\citep{openai_gpt5_system_card_2025}              & 75.22 & 0.1790 & 0.1535 & 0.1658 & 0.2114 & 0.1811 & 0.1957 \\
GPT-4o~\citep{hurst2024gpt}          & \textbf{{87.76}}  & \textbf{0.3750} & {0.3487} & {0.3612} & \textbf{0.4054} & {0.3770} & {0.3905} \\
\midrule
\rowcolor{gray!10}
AnomReasonor-7B        & 82.61 & 0.3684 & \textbf{0.3574} & \textbf{0.3613} & 0.4048 & 0.\textbf{3929} & \textbf{0.3972} \\

\bottomrule
\end{tabular}
}
\end{table}

\subsection{Explainable Deepfake Detection}
\label{subsec:deepfake}

We extend 
    our semantic anomaly reasoning framework (Sec.~\ref{subsec:main-results}) to an {explainable deepfake detection} setting. 
The goal is twofold: 
    (i) determine whether an image is AI-generated, 
    and (ii) return structured anomaly explanations aligned with Sec.~\ref{subsec:eval}. 
To this end, we construct \textbf{AnomReason-Deepfake}, 
    where real images sampled from LAION/reLAION-HR~\citep{laion_relaion_high_resolution_2023} are paired with content-based structured descriptions. 
The task therefore probes both AIGC detection and semantic explanation within a unified benchmark.

\textbf{Metrics.} 
In addition to 
    binary detection accuracy (Acc), we introduce {classification-aware} variants of our semantic metrics, 
    denoted as \(\mathrm{CSemAP}_v\) and \(\mathrm{CSemF1}_v\) for $v\in\{\mathrm{Phe},\mathrm{Rea},\mathrm{Full}\}$. 
Explanations 
    are scored only when detection is correct, and assigned zero otherwise. 
This ties explanation quality to valid classification, 
    discouraging post-hoc rationalization for mispredictions and promoting joint interpretability.

\textbf{Results.} 
Table~\ref{tab:deepfake_results} 
    presents results across Acc and classification-aware semantics metrics. 
GPT-4o achieves 
    the best overall detection accuracy (87.76\%) and strong \CSemFOneFull (0.3905), 
    establishing a strong proprietary baseline. 
Notably, our fine-tuned \textit{AnomReasonor-7B} 
    attains competitive accuracy (82.61\%) and \textbf{surpasses} GPT-4o on \CSemAPRea (0.3574 vs. 0.3487) and \CSemFOneRea (0.3929 vs. 0.3770), highlighting its strength in generating causally grounded explanations.
Open-source VLMs 
    show large variance. 
Larger models 
    like Qwen2.5-VL-72B achieve decent detection (77.60\%) and moderate explanation ability (\CSemFOneFull: 0.2266), but remain behind task-aligned models. 
Some models (e.g., InternVL3-8B) 
    perform reasonably on \CSemAPFull (0.2559) but poorly on F1, suggesting limited calibration between confidence and semantic consistency.
Overall, \textit{AnomReasonor-7B} offers a high-accuracy, 
    interpretable alternative to closed models. 
These results 
    underscore the utility of \textit{AnomReason-Deepfake} as a testbed for \textbf{joint detection and structured explanation}, and support our pipeline’s effectiveness for training explainable AIGC detectors.

\begin{table}[h]
\centering
\caption{Semantic Reasonableness of AIGC Image Generators.}
\label{tab:aigc_reasonableness}
\resizebox{0.9\textwidth}{!}{
\begin{tabular}{l ccc|ccc}
\toprule
\multirow{2}{*}{\textbf{AIGC Model}} & \multicolumn{3}{c}{\textbf{AnomReasonor}} & \multicolumn{3}{c}{\textbf{AnomAgent}} \\
\cmidrule(lr){2-7} & \textbf{MAI (\textdownarrow)} & \textbf{AF(\textdownarrow)} & \textbf{CAP(\textdownarrow)}  & \textbf{MAI (\textdownarrow)} & \textbf{AF(\textdownarrow)} & \textbf{CAP(\textdownarrow)}  \\
\midrule
Sana 1.5~\citep{xie2025sana}                              & 4.55 & 6.29 & 28.62 & 6.66 & 9.09  &  60.49\\
SDXL Lightning~\citep{lin2024sdxllightning}               & 4.40 & 6.06 & 26.66 & 6.47 & 8.84  &  57.21\\
Sana Sprint 1.6B~\citep{chen2025sanasprint}               & 4.32 & 6.09 & 26.31 & 6.42 & 8.78  &  56.35\\
Qwen-lmage~\citep{wu2025qwenimagetechnicalreport}         & 4.45 & 6.20 & 27.59 & 6.37 & 8.72  &  55.61\\
HiDream-1-Fast~\citep{hidreami1technicalreport}           & 4.44 & 6.19 & 27.48 & 6.32 & 8.81  &  55.68\\
SDv3.5 Large~\citep{esser2024scaling}                     & 4.60 & 6.44 & 29.62 & 6.23 & 8.54  &  53.19\\
Janus Pro 7B~\citep{chen2025janus}                        & 4.28 & 6.75 & 28.89 & 6.49 & 8.69  &  56.39\\
Janus Pro 1B~\citep{chen2025janus}                        & 4.19 & 6.43 & 26.94 & 6.40 & 8.41  &  53.83\\
FLUX.1 [dev]~\citep{wu2025qwenimagetechnicalreport}       & 4.36 & 6.08 & 26.51 & 6.22 & 8.67  &  53.92\\
BRIA-3.2    ~\citep{BriaAI_BRIA_3.2}                      & 4.38 & 6.14 & 26.89 & 6.11 & 8.60  &  52.61\\
SDv3.5 Large Turbo~\citep{esser2024scaling}               & 4.40 & 6.12 & 26.93 & 6.10 & 8.44  &  51.48\\
Lumina lmage V2~\citep{lumina2}                           & 4.30 & 6.05 & 26.01 & 6.07 & \textbf{8.40}  &  51.03\\
HiDream-1-Dev~\citep{hidreami1technicalreport}            & 4.42 & 6.15 & 27.18 & \textbf{6.00} & 8.44  &  \textbf{50.62}\\
OmniGen V2~\citep{wu2025omnigen2}                         & 4.21 & \textbf{5.89} & 24.80 & 6.11 & 8.49  &  51.86\\
HunyuanImage-2.1~\citep{HunyuanImage-2.1}                 & \textbf{4.10} & 5.95 & \textbf{24.40} & 6.14 & 8.55  &  52.51\\
\bottomrule
\end{tabular}
}
\end{table}

\subsection{AIGC Semantic Reasonableness Assessment}
\label{subsec:reasonableness}
Beyond 
    instance-level evaluation (Sec.~\ref{subsec:main-results}), we assess the {semantic reasonableness} of text-to-image generators. 
Perceptual metrics 
    overlook commonsense, physics, and interaction plausibility, whereas structured outputs (Name/Phenomenon/Reasoning/Severity Score) enable semantics-aware auditing at scale.

We curate 246 prompts 
    from Midjourney public galleries that correspond to real-photo style content. 
We first 
    deduplicate/cluster candidate prompts via CLIP embeddings, then filter to {photo-realistic} style using the Qwen3-30B~\citep{qwen3technicalreport}. 
This yields 
    diverse prompts covering multi-object interactions, human articulation, and physical dynamics.
Each generated image 
    is evaluated independently by two assessors: 
    (i) \textit{AnomReasonor-7B} (fine-tuned in Sec.~\ref{subsec:main-results}) and 
    (ii) \textit{AnomAgent}. 
Both produce anomalies with an Severity Score $s\in[0,100]$.

\textbf{Semantic Metrics.}
We define three complementary metrics to quantify the semantic plausibility of an image $I$:
(i) \textbf{MeanAnomalyImplausibility (MAI):}
    $
    \mathrm{MAI}(I) = \sum_{\hat{a} \in \widehat{\mathcal{A}}(I)} \frac{100 - s(\hat{a})}{100}
    $ 
    (ii)  \textbf{AnomalyFrequency (AF):} Total number of anomalies $\hat{a}$ identified in $I$.
    (iii)  \textbf{CumulativeAnomalyPenalty (CAP):}
    $
    \mathrm{CAP}(I) = \mathrm{MAI}(I) \times \mathrm{AF}(I)
    $
The $\mathrm{MAI}$ captures the aggregated implausibility of all detected anomalies. $\mathrm{CAP}$ reflects both the severity and frequency of semantic violations.
All scores are designed such that \textbf{lower is better}, with zero indicating flawless realism.

\textbf{Results.}
Table~\ref{tab:aigc_reasonableness} reveals clear differences in semantic plausibility across AIGC models. HunyuanImage-2.1 and OmniGen V2 achieve the lowest CAP scores under AnomReasonor, indicating fewer and less severe semantic anomalies, particularly in commonsense and physical interactions. In contrast, models like Sana 1.5 and SDXL Lightning exhibit higher anomaly frequency and implausibility, suggesting challenges in compositional reasoning and realism.
Interestingly, we observe distinct failure modes: some models (e.g., Janus Pro 7B) exhibit higher AF but lower MAI, implying frequent yet subtle errors, while others (e.g., SDv3.5 Large) show fewer but more severe implausibilities. These insights go beyond perceptual quality, revealing gaps in physical logic and social commonsense.
While both AnomReasonor and AnomAgent show consistent relative rankings, AnomAgent tends to produce higher anomaly counts across the board, reflecting its greater sensitivity and fully automated nature. Despite this, their strong rank-order alignment supports AnomAgent's robustness and suitability for scalable zero-shot auditing. Notably, AnomReasonor benefits from supervised training on human-curated explanations, whereas AnomAgent operates without any human intervention—highlighting the promise of fully automatic, semantics-aware evaluation pipelines for future alignment assessments.

\section{Related work}
\paragraph{AIGC visual anomaly detection and reasoning: }

With the proliferation of AI-generated content (AIGC), recent research~\citep{liu2024forgerygpt,huang2025sida,wen2025spot,kang2025legion,gao2025fakereasoning,huang2025so,zhang2025ivy,zhou2025aigi,tan2025veritas,guo2025rethinking,chen2024x2} has shifted beyond low-level forensics to explore semantic-level anomalies in AIGC images.
Some methods leverage VLM interpretability to highlight suspicious regions~\citep{liu2024forgerygpt,huang2025sida,kang2025legion}, while others combine manual inspection with LLM-based post-processing~\citep{li2025fakebench,tan2024ForenX}. Prompt-engineering and commonsense-injection techniques have also been used to elicit finer-grained descriptions~\citep{zhang2025ivy,gao2025fakereasoning,wen2025spot,zhou2025aigi}.

However, existing benchmarks such as FakeClue and Ivy‑Fake focus on authenticity classification or artifact explanation. They provide coarse labels or clues for real/fake discrimination, but lack structured quadruple annotations that capture commonsense, physical and relational inconsistencies.  Consequently, models trained on them struggle to perform detailed reasoning or severity assessment.
Our work differs by modelling anomalies at the object–attribute–relation level and offering interpretable explanations, thus enabling downstream tasks like semantic reasonableness auditing.  We also adopt a multi-agent annotation framework with human verification, which yields more consistent and scalable annotations than directly prompting a monolithic LLM.

\paragraph{AIGC Image Semantic Reasonableness Assessment:}

Assessing the \emph{semantic reasonableness} of AI-generated images (AIGC-ISRA) involves determining whether the visual content aligns with real-world commonsense, object plausibility, and coherent inter-object relationships.
While current AIGC image quality assessment efforts~\citep{peng2024aigc,liu2023g,lu2023llmscore,yang2024moe,yu2024sf} focuses primarily on image–prompt alignment and perceptual quality, but overlook semantic plausibility. 
For example, \citep{peng2024aigc}  propose CLIP-based metrics to assess prompt consistency, yet they fail to capture scene-level semantic inconsistencies.  
\citep{liu2023g} explore chain-of-thought evaluation for NLG tasks, offering improved human alignment but lacking grounding and localized reasoning for visual content.  
Our work addresses this gap via a structured and content-aware evaluation framework that builds upon a multi-agent annotator (\textit{AnomAgent}) to detect and explain attribute violations, spatial logic failures, and inter-object contradictions.  
This enables interpretable and fine-grained assessment of semantic plausibility, bridging a key limitation of existing AIGC-IQA and VLM-based approaches.

\section{Conclusion}

We introduce the task of \emph{semantic anomaly detection and reasoning} in AI‑generated image.  To enable this task, we built \textit{AnomReason}, a benchmark annotated via a multi-agent framework (\textit{AnomAgent}) and human verification, providing structured quadruples (Name, Phenomenon, Reasoning, Severity).  
Unlike existing authenticity datasets, AnomReason targets commonsense, physical and relational violations and includes severity scores, enabling finer-grained reasoning beyond real/fake classification.  
We propose semantics-aware metrics and showed that off-the-shelf vision-language models struggle on this task; a model fine-tuned on AnomReason (AnomReasonor-7B) outperformed open-source baselines and approached proprietary systems.  We also demonstrated applications in explainable deepfake detection and generator assessment.  
Current limitations include the dataset’s moderate scale and focus on images; future work will extend to videos and refine annotation quality.  
We will release code, data and models to foster research into semantic plausibility and more trustworthy multimodal systems.

\bibliographystyle{unsrt}  

\bibliography{references}


\newpage

\appendix

\begin{figure}[htp!]
    \centering
    \includegraphics[width=0.9\linewidth]{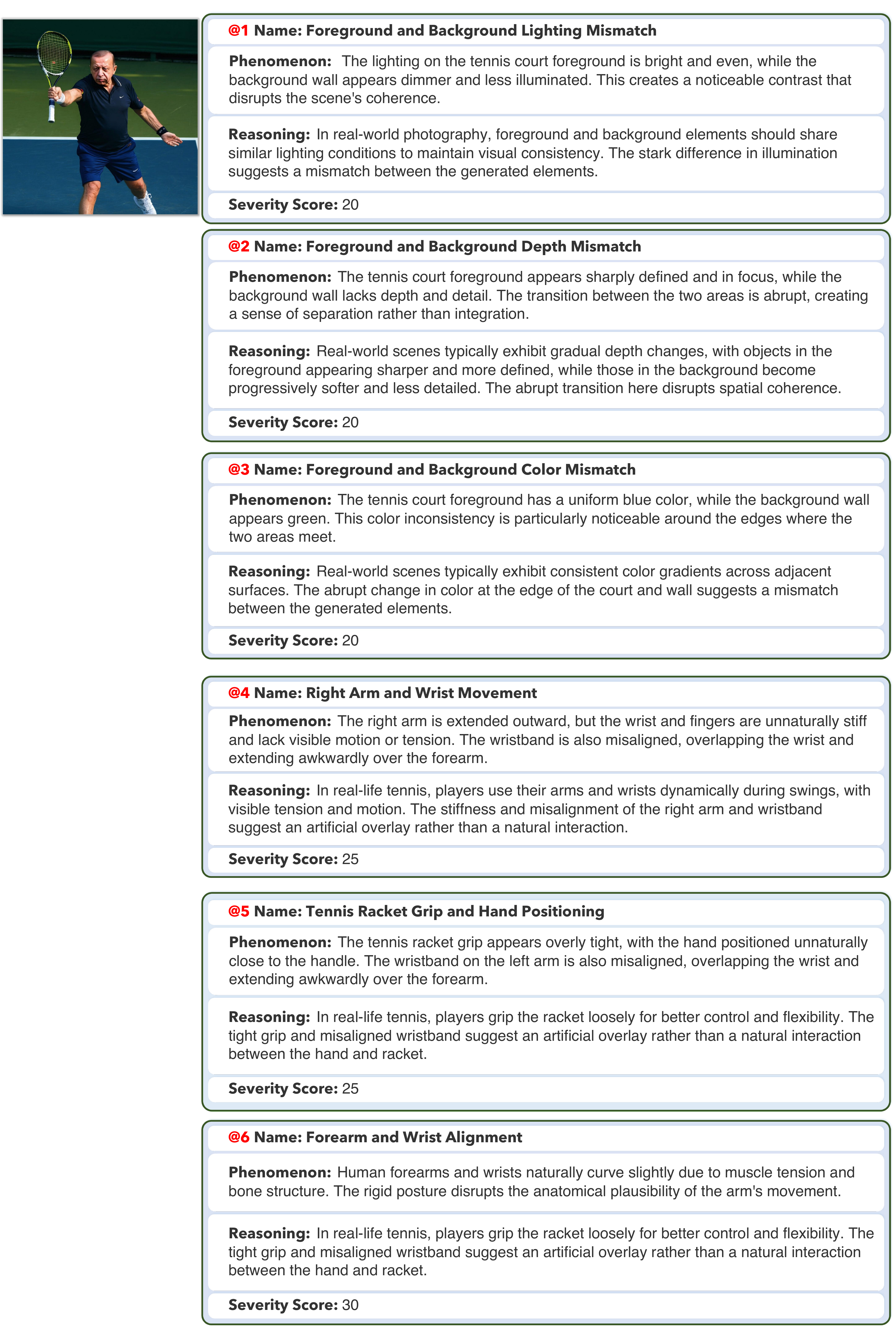}
    \caption{
    \textbf{Structured visual anomalies in a tennis scene.} 
    AnomReasonor-7B identifies both surface-level inconsistencies (e.g., lighting and color mismatch) and deeper semantic-level anomalies, such as biomechanically implausible wrist articulation and unnatural hand–racket interaction. Each anomaly is described with a structured triplet: \texttt{Name}, \texttt{Phenomenon}, \texttt{Reasoning}.
    }
    \label{fig:DRAnomReasonor_results_1}
\end{figure}

\begin{figure}[htp!]
    \centering
    \includegraphics[width=0.9\linewidth]{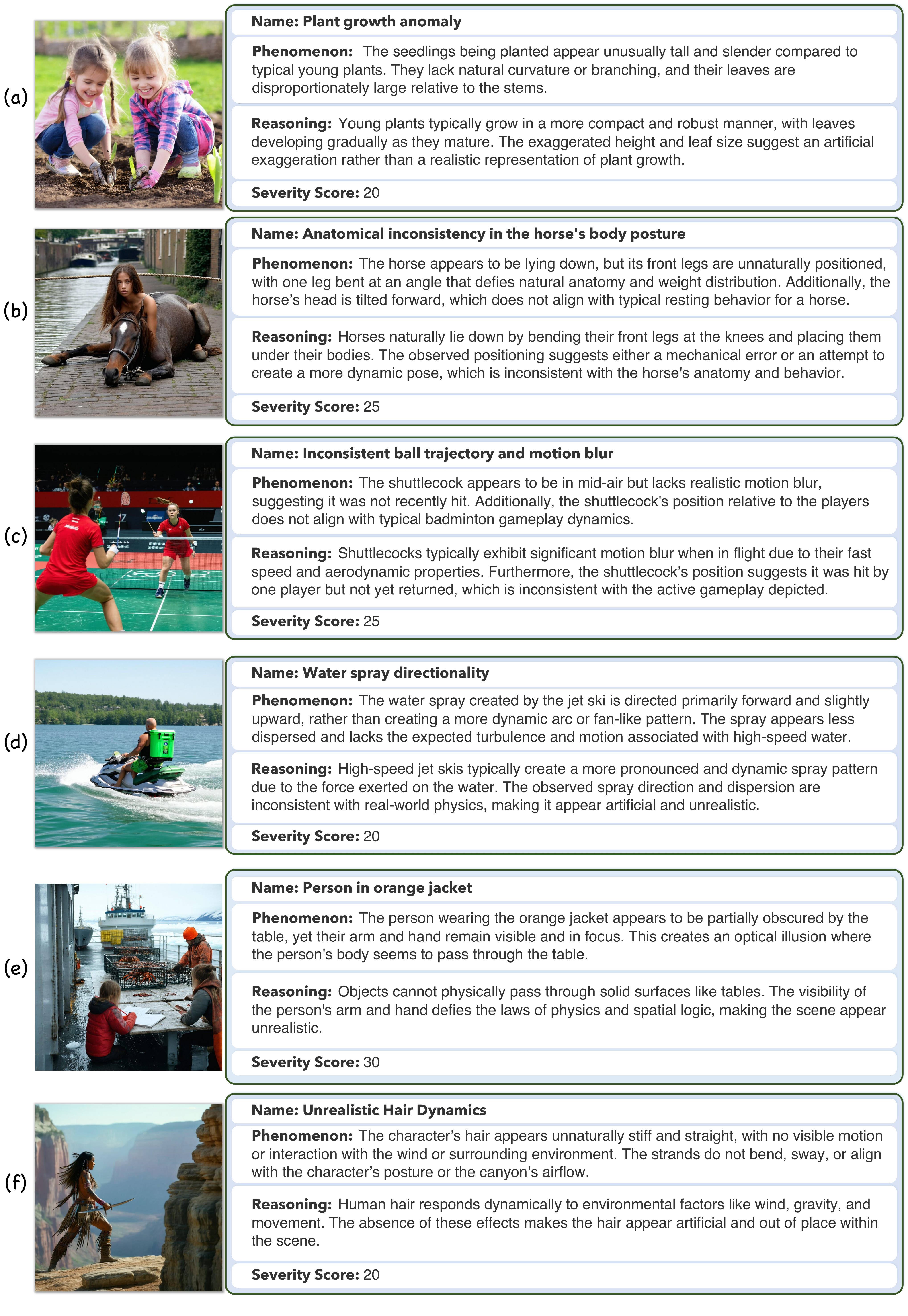}
    \caption{
    \textbf{Generalization across diverse semantic anomaly types.}
    AnomReasonor-7B detects inconsistencies in plant growth patterns (a), horse anatomy (b), shuttlecock dynamics (c), water spray physics (d), occlusion logic (e), and hair movement (f). These outputs demonstrate the model’s capability to reason over visual semantics beyond surface-level cues.
    }
    \label{fig:DRAnomReasonor_results_2}
\end{figure}

\section{Detection Results of AnomReasonor}
\label{appendix:DRAnomReasonor}

We present qualitative results from \textit{AnomReasonor-7B}, trained on our proposed \textit{AnomReason} dataset, to analyze its ability to detect and explain {semantic-level visual anomalies} in AI-generated images. Unlike {surface-level cues}—such as pixel artifacts, texture mismatches, or inconsistent lighting—{semantic-level anomalies} reflect inconsistencies with real-world physical laws, commonsense reasoning, spatial logic, and anatomical plausibility.

Figures~\ref{fig:DRAnomReasonor_results_1}–\ref{fig:DRAnomReasonor_results_2} highlight how structured reasoning outputs—composed of \texttt{Name}, \texttt{Phenomenon}, \texttt{Reasoning}, and \texttt{Severity Score}—enable fine-grained interpretation of visual errors that would be challenging to localize using traditional classification-based or pixel-level approaches.

Figure~\ref{fig:DRAnomReasonor_results_1} presents a synthesized tennis court scene containing a variety of both surface- and semantic-level anomalies. The model detects lighting, depth, and color mismatches between foreground and background, indicating surface inconsistencies. More critically, it identifies human-centric semantic anomalies including biomechanically implausible wrist articulation, unnatural racket grip tension, and anatomically misaligned arm–forearm joints. These violations of physical and anatomical realism are explicitly explained, demonstrating the model’s understanding of dynamic human interaction and contextual coherence.

Figure~\ref{fig:DRAnomReasonor_results_2} demonstrates the model’s ability to generalize across diverse contexts and anomaly types. Examples include:
\begin{itemize}
    \item[] \textbf{(a)} exaggerated plant growth violating botanical development norms,
    \item[] \textbf{(b)} a horse in an anatomically impossible lying posture,
    \item[] \textbf{(c)} a shuttlecock trajectory inconsistent with typical gameplay physics,
    \item[] \textbf{(d)} jet ski spray patterns that defy real-world water dynamics,
    \item[] \textbf{(e)} human limbs passing through occluding surfaces,
    \item[] \textbf{(f)} static hair ignoring environmental effects such as gravity or wind.
\end{itemize}
These structured explanations go beyond low-level discrepancies, offering human-aligned insight into the visual implausibility of each scene.

These results showcase the benefit of \textit{semantic-level anomaly detection} for building interpretable and trustworthy AIGC assessment systems. By explicitly modeling and articulating inconsistencies in visual semantics—rather than relying on post-hoc explanations or pixel cues—\textit{AnomReasonor-7B} provides rich, localized, and human-aligned insights into failure modes in AI-generated images.

% \appendix
\section{Evaluation Protocol Details}
\label{appendix:sem-metrics}

This appendix provides a comprehensive description of our semantic evaluation protocol, used to score structured visual anomaly predictions as described in Sec.~\ref{subsec:eval}.

\subsection{Semantic similarity: BERTScore configuration}
\label{app:bert}
We instantiate all field-wise similarities with \emph{BERTScore} (F1 variant). Unless noted, we freeze the backbone and preprocessing across all experiments to ensure comparability.
Given hypothesis $h$ and reference $r$, we use the BERTScore F1:
\begin{equation}
\mathrm{BERTScore}(h,r)=\frac{2\,P(h,r)\,R(h,r)}{P(h,r)+R(h,r)}\in[0,1],
\end{equation}
and define
\begin{align*}
\mathrm{sim}_{\mathrm{Phe}} &=\mathrm{BERTScore}(\hat o,o),\\
\mathrm{sim}_{\mathrm{Rea}} &=\mathrm{BERTScore}(\hat r,r),\\
\mathrm{sim}_{\mathrm{Full}} &= \alpha~\mathrm{sim}_{\mathrm{Phe}}+(1-\alpha)\mathrm{sim}_{\mathrm{Rea}}.
\end{align*}
The balanced mean for \textbf{Full} provides a smooth joint signal while discouraging over-optimizing a single field. We use $\alpha = 0.5$ unless otherwise stated.

\subsection{Matching and contingency counts}
\label{app:matching}
For a view $v\in\{\mathrm{Phe},\mathrm{Rea},\mathrm{Full}\}$ and threshold $\tau\in\{0.7,0.8,0.9\}$, define the indicator
\begin{equation}
T_v(\hat a,a;\tau)=\mathbb{1}\{\mathrm{sim}_v(\hat a,a)\ge\tau\}.
\end{equation}
\textbf{Ranking.} Sort predictions $\widehat{\mathcal{A}}(I)=\{\hat a_{(k)}\}_{k=1}^{K_I}$ by confidence $\hat s$ in descending order.  
\textbf{Assignment.} Scan $k=1\ldots K_I$; for each $\hat a_{(k)}$ choose the unmatched ground-truth $a^\star\in\mathcal{A}(I)$ maximizing $\mathrm{sim}_v(\hat a_{(k)},a)$ subject to $T_v=1$. If such $a^\star$ exists, mark a true positive (TP) and lock both; else, mark a false positive (FP). Unmatched ground truths are false negatives (FN).  
\textbf{Tie-breaking.} If multiple $a$ achieve the same similarity, prefer the one with the highest $\mathrm{sim}_{\mathrm{Full}}$ (then smallest index).  
This one-to-one greedy rule prevents multi-matching and implicitly penalizes duplicates.

\subsection{Per-image AP/F1, per-threshold aggregation, dataset aggregation}
\label{app:aggregation}
For image $I$ at threshold $\tau$, let cumulative counts at rank $k$ be $\mathrm{TP}_I(k,\tau)$ and $\mathrm{FP}_I(k,\tau)$. Define
\begin{equation}
P_I(k,\tau)=\frac{\mathrm{TP}_I(k,\tau)}{\mathrm{TP}_I(k,\tau)+\mathrm{FP}_I(k,\tau)},\qquad
R_I(k,\tau)=\frac{\mathrm{TP}_I(k,\tau)}{|\mathcal{A}(I)|}.
\end{equation}
\textbf{AP at $\tau$.}
\begin{equation}
\mathrm{AP}_v(I,\tau)=\!\!\sum_{k:\,\text{new TP at }k}\! P_I(k,\tau)\,\big(R_I(k,\tau)-R_I(k\!-\!1,\tau)\big).
\end{equation}
\textbf{Per-image SemAP (average over thresholds).}
\begin{equation}
\mathrm{SemAP}_v(I)=\frac{1}{|\mathcal{T}|}\sum_{\tau\in\mathcal{T}}\mathrm{AP}_v(I,\tau),\quad \mathcal{T}=\{0.7,0.8,0.9\},\ |\mathcal{T}|=3.
\end{equation}
\textbf{Dataset SemAP (macro over images).}
\begin{equation}
\mathrm{SemAP}_v=\frac{1}{|\mathcal{D}|}\sum_{I\in\mathcal{D}}\mathrm{SemAP}_v(I).
\end{equation}
\textbf{Per-image F1 at $\tau$.}
\begin{equation}
\mathrm{F1}_v(I,\tau)=\frac{2\,P_I(\tau)\,R_I(\tau)}{P_I(\tau)+R_I(\tau)},\quad 
P_I(\tau)=\frac{\mathrm{TP}_I}{\mathrm{TP}_I+\mathrm{FP}_I},\ R_I(\tau)=\frac{\mathrm{TP}_I}{|\mathcal{A}(I)|}.
\end{equation}
\textbf{Dataset SemF1.} We report the simple average across images and thresholds:
\begin{equation}
\mathrm{SemF1}_v=\frac{1}{|\mathcal{D}|\,|\mathcal{T}|}\sum_{I\in\mathcal{D}}\sum_{\tau\in\mathcal{T}}\mathrm{F1}_v(I,\tau),
\end{equation}
with per-threshold breakdowns provided below.

In Sec. \ref{subsec:deepfake}, we introduce \emph{classification-aware} variants of our semantic metrics, denoted as \(\mathrm{CSemAP}_v\) and \(\mathrm{CSemF1}_v\) for $v\in\{\mathrm{Phe},\mathrm{Rea},\mathrm{Full}\}$. 
Let $y(I)\in\{\text{real},\text{AI}\}$ be the ground-truth source and $\hat y(I)$ the predicted source. For a view $v\in\{\mathrm{Phe},\mathrm{Rea},\mathrm{Full}\}$ and threshold set $\mathcal{T}=\{0.7,0.8,0.9\}$, define
\begin{equation}
\mathrm{CSemAP}_v \;=\; \frac{1}{|\mathcal{D}|\,|\mathcal{T}|} \sum_{I\in\mathcal{D}}\ \sum_{\tau\in\mathcal{T}} 
\mathbb{1}\{\hat y(I)=y(I)\}\;\mathrm{AP}_v(I,\tau),
\end{equation}
\begin{equation}
\mathrm{CSemF1}_v \;=\; \frac{1}{|\mathcal{D}|\,|\mathcal{T}|} \sum_{I\in\mathcal{D}}\ \sum_{\tau\in\mathcal{T}} 
\mathbb{1}\{\hat y(I)=y(I)\}\;\mathrm{F1}_v(I,\tau),
\end{equation}
Explanations are scored only when detection is correct, and assigned zero otherwise. This ties explanation quality to valid classification, discouraging post-hoc rationalization for mispredictions and promoting joint interpretability.

\section{Training Details}
\label{app:training}

\subsection{Details for Sec.\ref{subsec:main-results}}
\label{app:training:model}
We fine-tune Qwen2.5-VL-7B using LoRA modules inserted into the multimodal attention blocks, while keeping the visual encoder frozen. Specifically, we adopt the official Qwen2.5-VL-7B backbone (7B parameters) with its native image projector, freezing the vision tower (ViT). LoRA is applied to the attention projections (\texttt{q\_proj}, \texttt{k\_proj}, \texttt{v\_proj}) at every 4th transformer layer, with a rank of 8, scaling factor $\alpha = 16$, and dropout rate of 0.5. Fine-tuning is conducted for one epoch using a batch size of 4 and gradient accumulation of 8 on 4 A6000 GPUs. Input resolution is dynamically adjusted with a minimum of 50,176 pixels and a maximum of 614,656 pixels.

\paragraph{Supervision data and construction}
We use the official \texttt{train} split of the \textit{AnomReason} dataset for training, strictly retaining only \emph{verified} anomalies to ensure label quality. For each image \( I \), we construct one or multiple training instances depending on sequence length constraints. When an image contains \( m \) verified anomalies \(\{(y_j, o_j, r_j, v_j)\}_{j=1}^{m}\), we pack them into a single instance in randomized order unless token limits require splitting.

\begin{quote}
\ttfamily
Human: Analyze the provided AI-generated image to detect all semantic anomalies. Provide a detailed list of anomalies using the following structure:  
- @1. \textbf{Name}: [Phenomenon Name]  
- \textbf{Phenomenon}: [Description of the visible anomaly]  
- \textbf{Reasoning}: [Step-by-step explanation of why this phenomenon is unrealistic or illogical]  
- \textbf{Severity Score}: [Rate the issue from 0 (completely unnatural) to 100 (fully realistic)]
\end{quote}

The model is expected to produce responses such as:

\begin{quote}
\ttfamily
@1. \textbf{Name}: Woman’s shirt fabric rigidity\\
\textbf{Phenomenon}: The shirt worn by the woman appears overly smooth and rigid, showing no folds, creases, or deformation where it interacts with the baby or the reclining posture.\\
\textbf{Reasoning}: Soft fabrics typically conform to the body and surrounding objects, creating visible folds and compression due to gravity and physical contact. The lack of such effects reduces the plausibility of the material behavior.\\
\textbf{Severity Score}: 20.\\

@2. \textbf{Name}: Pillow compression inconsistency\\
\textbf{Phenomenon}: The pillow supporting the woman’s head and shoulder appears unnaturally smooth and rigid, showing no visible deformation or compression despite the weight applied.\\
\textbf{Reasoning}: Pillows are designed to compress under pressure, creating indentations and folds where weight is applied. The lack of deformation suggests unrealistic material behavior, detracting from the plausibility of the scene.\\
\textbf{Severity Score}: 25.
\end{quote}

This format ensures consistent alignment between training supervision and evaluation, facilitating learning of structured anomaly descriptions and reasoning patterns. The use of step-wise justifications and severity scoring further supports interpretable and rank-aware detection.

\subsection{Details for Sec.\ref{subsec:deepfake}}

We fine-tune \textbf{Qwen2.5-VL-7B} using low-rank adapters (LoRA), following the same hyperparameters described in Appendix~\ref{app:training:model}, with the exception that we set the dropout rate for LoRA (\texttt{lora\_dropout}) to 0.9 for this setting.

\paragraph{Turn 1: Source classification.}
The first turn asks the model to determine the image source in binary form:

\begin{quote}
\ttfamily
Q: Summarize whether this image is Generated by Artificial Intelligence, please only return yes or no.\\
A: \textless Yes, this image is generated by AI | No, this image is a real photograph.\textgreater
\end{quote}

\paragraph{Turn 2: Semantic anomaly explanation.}
If the image is AI-generated, the second turn prompts the model to enumerate all semantic-level inconsistencies using a structured output format:

\begin{quote}
\ttfamily
Q: If AI-generated, list semantic anomalies:\\
\quad Name, Phenomenon, Reasoning, Severity.\\
A: [\\
\quad \{Name: \textless y1\textgreater, Observed: \textless o1\textgreater, Reasoning: \textless r1\textgreater, Severity: \textless v1\textgreater\},\\
\quad \ldots\\
]
\end{quote}

If the image is labeled as {real}, the second answer defaults to a content-based explanation of what is observed in the image without marking any anomaly. This dual-turn QA formulation enables the model to explicitly separate generation source classification from semantic inconsistency identification, promoting robust deepfake understanding and explainability.

\section{Additional Details on Anomaly Detection Agents}
\label{sec:agent_details}

In this section, we provide detailed insights into the design and implementation of the anomaly detection agents used in our framework. Each agent plays a crucial role in analyzing different aspects of the input image, allowing for an in-depth semantic analysis of AI-generated images. The design of these agents is driven by the need to model high-level perceptual reasoning and commonsense logic, which are essential for detecting complex semantic anomalies that would otherwise go unnoticed by traditional anomaly detection methods.

\subsection{Object Perceiver Agent}

The \textbf{ObjectPerceiver} agent is tasked with identifying and parsing all semantically distinct entities within an input image. This is the first step in the anomaly detection pipeline, as it helps isolate the relevant objects in the image for further analysis. By detecting objects in the image, the agent serves as the foundation for all subsequent analysis.

\noindent \textbf{Design Motivation:}
The motivation behind the Object Perceiver is to provide a structured identification of objects in the image, which can then be used for deeper anomaly detection in subsequent steps. The design intentionally focuses on high-level object semantics rather than low-level features like lighting or texture inconsistencies, which might not be as impactful for detecting semantic anomalies.

\noindent
\begin{quote}
\ttfamily
\textbf{Task:} Analyze all objects and individuals in the image. For each object or individual, provide a detailed, accurate, and comprehensive description, while identifying any inconsistencies, anomalies, or illogical aspects. Ensure no object or body part is omitted.

\vspace{1em}
\textbf{Follow the steps below and provide your analysis in the structured format specified:}
\begin{itemize}
    \item Identify and describe all objects and individuals in the image.
    \item For each object or individual, provide a detailed, accurate, and comprehensive description.
    \item Highlight any inconsistencies, anomalies, or illogical aspects in:
    \begin{itemize}
        \item \textbf{Shape and Structure}: Are there distortions, missing parts, or unnatural forms?
        \item \textbf{Material and Texture}: Are there abrupt texture changes or mismatches?
        \item \textbf{Lighting and Shadows}: Are the lighting and shadows consistent with the environment?
        \item \textbf{Physical Properties}: Are there any violations of real-world physics or logic (e.g., floating objects)?
        \item \textbf{Common Sense Verification}: Are there any semantic inconsistencies (e.g., a door handle on a chair)?
        \item \textbf{Human Anatomy (if applicable)}: Identify unnatural features such as missing limbs, extra fingers, or disproportionate body parts.
    \end{itemize}
\end{itemize}

\vspace{1em}
\textbf{Output Format:} \\
Each object/body part should be described individually in the following structured format:

\begin{verbatim}
#Name#: Detailed Description.

#Name#: Detailed Description.
\end{verbatim}

\textbf{Example Output:}
\begin{itemize}[leftmargin=-0.05em]
\item[] Person: A man with three arms, one of which is unnaturally attached to his back. He wears a blue jacket.
\item[] Chair: A wooden chair that appears to be floating without support, casting no shadow.
\item[] Dog: A golden retriever with two tails, one of which is blurry and semi-transparent.
\end{itemize}

Highlight all implausible, unnatural, or inconsistent details while ensuring full coverage of the image content. Only output the list in the specified format.
\end{quote}

\noindent \textbf{Effectiveness:}
This prompt ensures that the Object Perceiver focuses on capturing the main objects and entities in the image while disregarding irrelevant details. By emphasizing the semantics of each object, the agent avoids common pitfalls of pixel-based anomaly detection and lays the groundwork for higher-level analysis.

\subsection{Attribute Analyzer Agent}

The \textbf{Attribute Analyzer} is responsible for detecting anomalies in the visual attributes of objects, such as shape, texture, material, and other intrinsic properties. This agent examines internal inconsistencies within the objects themselves and identifies any attributes that deviate from the expected real-world norms.

\noindent \textbf{Design Motivation:}
The rationale behind the Attribute Analyzer is that many semantic anomalies manifest as inconsistencies in object attributes. For example, an object might appear out of place because of an unusual shape or texture that violates expectations based on the scene context. By focusing on attributes, this agent can detect low-level anomalies that might not be caught by higher-level reasoning alone.

\noindent \textbf{AttributeAnalyzer Step1 Prompt:}
\begin{quote}
\ttfamily
\textbf{Task}: Analyze \textbf{\{Current object\}} in the image.\\

Focus on analyzing and identifying any anomalies in the following aspects:

\begin{enumerate}[leftmargin=2.5em]
    \item \textbf{Shape and Structure}
    \begin{itemize}
        \item Are there unnatural forms or distortions?
        \item Are proportions consistent with the object's design?
    \end{itemize}

    \item \textbf{Functionality}
    \begin{itemize}
        \item Does the object behave logically in real-world scenarios?
        \item Are there physical impossibilities (e.g., unsupported structures)?
    \end{itemize}

    \item \textbf{Human Body Structure Verification} (if applicable)
    \begin{itemize}
        \item Are limbs, fingers, and facial features correctly placed and proportional?
        \item Are there unnatural fusions, duplications, or disconnections?
    \end{itemize}
\end{enumerate}

\vspace{0.5em}
\textbf{Deliverable}:
\begin{itemize}[leftmargin=2.5em]
    \item Highlight all implausible, unnatural, or inconsistent details.
    \item Ensure a thorough analysis that covers all aspects of the image content.
    \item Provide concise, evidence-based explanations for all findings.
\end{itemize}
\end{quote}

\noindent \textbf{AttributeAnalyzer Step2 Prompt:}

\begin{quote}
\ttfamily
\noindent\textbf{Object:} \textbf{\{Current object\}} \\
\textbf{Description Input:} \texttt{\{AttributeAnalyzer Step1 Response\}}

\vspace{1em}

\noindent\textbf{Task}: Analyze the detailed description of \textbf{\{Current object\}} and identify all unreasonable, contradictory, or physically impossible details specific to this object.

\vspace{0.5em}
\noindent\textbf{Provide a structured list of issues using the following format:}
\begin{itemize}[leftmargin=2.5em]
    \item \textbf{Abnormal Phenomenon Name}: The name of the observed anomaly.
    \item \textbf{Observed Issue}: The unnatural feature found.
    \item \textbf{Explanation}: Why this characteristic is unrealistic.
\end{itemize}

\vspace{0.5em}
\noindent\textbf{Example Output:}
\begin{enumerate}[leftmargin=2.5em]
    \item \textbf{Abnormal Phenomenon Name}: Streetlight No Power \\
    \textbf{Observed Issue}: The streetlight is glowing but has no power source or wiring. \\
    \textbf{Explanation}: A streetlight requires an electrical connection to function, and no wires or batteries are visible.
\end{enumerate}

\vspace{0.5em}
\noindent\textbf{Instructions:}
\begin{itemize}[leftmargin=2.5em]
    \item Analyze \textbf{only} \{Current object\}.
    \item Output \textbf{only} issues directly related to \{Current object\}, using the specified format.
\end{itemize}
\end{quote}

\noindent \textbf{Effectiveness:}
This two-stage approach encourages detailed yet structured interpretation of each object’s attributes. The separation of observation and reasoning mimics human perceptual processes and aligns well with structured evaluation protocols introduced in Sec.~\ref{subsec:eval}.

\subsection{Relation Reasoner Agent}

The \textbf{RelationReasoner} analyzes spatial, semantic, and functional relationships between objects in the image. It checks if objects are interacting in a plausible way or if their relationships defy physical or logical laws. This agent is critical for detecting anomalies that arise from the interaction between objects.

\noindent \textbf{Design Motivation:}
The design of the Relation Reasoner is motivated by the fact that many semantic anomalies stem from the way objects relate to one another. For example, objects that should interact, such as a hand holding a cup, may appear to be floating or not touching each other at all. By modeling relationships, the agent detects high-level semantic inconsistencies that are often missed by attribute analysis alone.

\noindent \textbf{RelationReasoner Step1 Prompt:}
\begin{quote}
\ttfamily
\noindent\textbf{Task}: Analyze the spatial and logical relationships between \textbf{\{Current object\}} and the following objects: (\textit{\{all objects from ObjectPerceiver\}}). 

You should evaluate:
\begin{itemize}[leftmargin=2em]
    \item One-to-one relationships (e.g., \textbf{\{Current object\}} with each object)
    \item One-to-many relationships (e.g., \textbf{\{Current object\}} in relation to multiple objects collectively)
\end{itemize}

\vspace{0.5em}
\noindent\textbf{Context Descriptions}: \\
\texttt{\{Object Descriptions from AttributeAnalyzer\}}

\vspace{1em}
\noindent\textbf{Focus Areas}:
\begin{enumerate}[leftmargin=2em]
    \item \textbf{Perspective Errors}: Are objects placed in impossible or illogical locations relative to \textbf{\{Current object\}}?
    \item \textbf{Physical Interactions}: Are there unnatural interactions (e.g., floating without support, overlapping unnaturally)?
    \item \textbf{Logical Contradictions}: Are there contradictions with real-world behavior or common-sense logic?
\end{enumerate}

\vspace{1em}
\noindent\textbf{Instructions}:
\begin{itemize}[leftmargin=2em]
    \item Focus analysis on \textbf{\{Current object\}} as the primary subject.
    \item For \textbf{one-to-one relationships}, evaluate individual pairings.
    \item For \textbf{one-to-many relationships}, consider collective spatial, logical, and contextual coherence.
\end{itemize}

\vspace{1em}
\noindent\textbf{Output Format} (structured report for each issue):
\begin{itemize}[leftmargin=2.5em]
    \item \textbf{Relationship}: Describe the relationship being analyzed.
    \item \textbf{Observed Issue}: Detail the anomaly or inconsistency.
    \item \textbf{Explanation}: Explain why the issue is illogical or unrealistic.
\end{itemize}

\vspace{0.5em}
\noindent\textbf{Deliverables}:
\begin{itemize}[leftmargin=2em]
    \item Analyze all one-to-one and one-to-many relationships involving \textbf{\{Current object\}}.
    \item Ensure detailed reasoning and structured output for each detected issue.
\end{itemize}
\end{quote}

\noindent \textbf{RelationReasoner Step2 Prompt:}
\begin{quote}
\ttfamily
\noindent\textbf{Relation Input:} \texttt{\{RelationReasoner Step1 Response\}}

\vspace{1em}
\noindent\textbf{Focus Object}: The primary subject of analysis is \textbf{\{Current object\}}. All evaluations should center on \textbf{\{Current object\}} and its relationships with the following objects: (\textit{\{all objects from ObjectPerceiver\}}).

\vspace{0.5em}
\noindent\textbf{Task}: Based on the prior relationship analysis, analyze and summarize the relationships between \textbf{\{Current object\}} and the listed objects. Emphasize detection of logical contradictions, physical impossibilities, and semantic anomalies.

\vspace{1em}
\noindent\textbf{Key Aspects to Evaluate}:
\begin{enumerate}[leftmargin=2em]
    \item \textbf{Logical Coherence}: Are the relationships internally consistent?
    \begin{itemize}
        \item Example: An object cannot be both inside and outside another simultaneously.
    \end{itemize}

    \item \textbf{Physical Realism}: Do the relationships conform to real-world physical laws?
    \begin{itemize}
        \item Example: Objects should not float without visible support.
    \end{itemize}

    \item \textbf{Semantic Plausibility}: Are the interactions meaningful and contextually appropriate?
    \begin{itemize}
        \item Example: A dog “wearing” a cloud is not semantically plausible.
    \end{itemize}

    \item \textbf{Causal Consistency}: Do object states logically follow from their relationships?
    \begin{itemize}
        \item Example: A book balanced on a steep slope should be expected to fall.
    \end{itemize}
\end{enumerate}

\vspace{1em}
\noindent\textbf{Output Format}: For each detected anomaly, provide a structured report as follows:
\begin{itemize}[leftmargin=2.5em]
    \item \textbf{Objects Involved}: List the relevant objects (including \textbf{\{Current object\}}).
    \item \textbf{Observed Issue}: Describe the logical, physical, or semantic anomaly.
    \item \textbf{Reasoning}: Justify why this relationship is unnatural, implausible, or illogical.
\end{itemize}

\vspace{0.5em}
\noindent\textbf{Instructions}:
\begin{itemize}[leftmargin=2em]
    \item Focus exclusively on \textbf{\{Current object\}} and its relationships.
    \item Evaluate both individual (one-to-one) and group (one-to-many) relationships.
\end{itemize}
\end{quote}

\noindent \textbf{Effectiveness:}
By evaluating the spatial and functional relations between objects, the Relation Reasoner helps detect anomalies that would violate everyday common sense or physical laws. For example, it would catch a cup floating in mid-air without any support or a person walking through a solid wall. This agent significantly enhances the ability to detect high-level semantic anomalies that go beyond object-level properties.

\subsection{Anomaly Integration and Formatting}

After the anomalies are detected by the Attribute Analyzer and Relation Reasoner, they are passed to the \textbf{Anomaly Integration} and \textbf{Anomaly Formatting} agents. The Integration agent consolidates similar or redundant anomalies and eliminates noise, while the Formatting agent structures the anomalies into a final output format.

\noindent \textbf{Design Motivation:}
The design of the Anomaly Integration and Formatting agents is driven by the need to ensure that the detected anomalies are presented in a clear and interpretable manner. The Integration agent helps combine similar anomalies into one, while the Formatting agent ensures the final output is structured and easy to use for downstream applications like quality assessment or dataset debugging.

\noindent \textbf{ AnomalyIntegrator Step1 Prompt:}
\begin{quote}
\ttfamily
\noindent Description for \textbf{\{Current object\}}: \texttt{\{AttributeAnalyzer Response\}} \\
Relationships of \textbf{\{Current object\}} with other objects (\textbf{\{all objects from ObjectPerceiver\}}): \texttt{\{RelationReasoner Response\}}

\vspace{1em}
\noindent\textbf{Task}: Review and analyze the detailed \textbf{Description} and \textbf{Relationships} of \textbf{\{Current object\}}. Summarize all unreasonable, contradictory, or physically impossible details related to \textbf{\{Current object\}}, while consolidating similar or repeated anomalies into a comprehensive report.

\vspace{0.5em}
\noindent\textbf{Focus Areas}:
\begin{enumerate}[leftmargin=2em]
    \item \textbf{Contradictory Details}: Identify conflicting statements or relationships (e.g., "floating" vs. "resting on the ground").
    \item \textbf{Unnatural Behaviors}: Highlight features or actions implausible in real-world settings.
    \item \textbf{Spatial Inconsistencies}: Detect impossible locations or orientations for \textbf{\{Current object\}} or others.
    \item \textbf{Illogical Physical Properties}: Point out violations of physics or reality (e.g., water flowing upward).
\end{enumerate}

\vspace{0.5em}
\noindent\textbf{Instructions}:
\begin{itemize}[leftmargin=2em]
    \item Consolidate similar anomalies from both \textbf{Description} and \textbf{Relationships}.
    \item Center all findings around \textbf{\{Current object\}} and its interactions with other objects.
\end{itemize}

\vspace{0.5em}
\noindent\textbf{Output Format} (structured list):
\begin{enumerate}[leftmargin=2em]
    \item \textbf{Observed Phenomenon}: \textit{Brief summary of the inconsistency}
    \begin{itemize}
        \item \textbf{Sources}: Indicate if the issue comes from the \textbf{Description}, \textbf{Relationships}, or \textbf{Both}.
        \item \textbf{Details}: Provide specific observations related to the anomaly.
        \item \textbf{Explanation}: Justify why the phenomenon is contradictory, unnatural, or implausible.
    \end{itemize}
\end{enumerate}

\vspace{0.5em}
\noindent\textbf{Deliverables}:
\begin{itemize}[leftmargin=2em]
    \item Focus exclusively on \textbf{\{Current object\}}.
    \item Consolidate and summarize issues across \textbf{Description} and \textbf{Relationships}.
    \item Output only the structured list in the specified format.
\end{itemize}
\end{quote}

\noindent \textbf{AnomalyIntegrator Step2 Prompt:}
\begin{quote}
\ttfamily
\noindent\textbf{Anomalies:} \texttt{\{AnomalyIntegrator Step1 Response\}}

\vspace{1em}
\noindent\textbf{Task}: Summarize and categorize all detected unnatural, illogical, or inconsistent phenomena in the image.

\vspace{0.5em}
\noindent\textbf{For each issue, provide:}
\begin{enumerate}[leftmargin=2em]
    \item \textbf{Object Name}: Clearly identify the object(s) involved.
    \item \textbf{Phenomenon}: Describe the unnatural or illogical aspect of the object(s).
    \item \textbf{Explanation}: Explain why this phenomenon is unrealistic, referencing real-world physics, anatomy, perspective, or common sense.
\end{enumerate}

\vspace{1em}
\noindent\textbf{Output Format}:
Provide a structured list using the format below:

\vspace{0.5em}
\noindent\textbf{Example Output:}
\begin{enumerate}[leftmargin=2em]
    \item \textbf{Object Name}: Tree \\
    \textbf{Phenomenon}: The tree trunk bends at an impossible 90-degree angle. \\
    \textbf{Explanation}: Real trees cannot grow in this shape due to gravitational constraints.

    \item \textbf{Object Name}: Dog \\
    \textbf{Phenomenon}: The dog has three tails, one of which is semi-transparent. \\
    \textbf{Explanation}: This is anatomically impossible for dogs.
\end{enumerate}

\vspace{0.5em}
\noindent\textbf{Instructions}:
\begin{itemize}[leftmargin=2em]
    \item Only output the list in the specified format.
    \item Ensure each anomaly is clearly tied to a specific object.
    \item Exclude unrelated content or commentary.
\end{itemize}
\end{quote}

\noindent \textbf{AnomalyFormatter Prompt:}
\begin{quote}
\ttfamily
\noindent\textbf{The following are a list of pre-selected anomalies:} \\
\texttt{\{List of AnomalyIntegrator Step2 Response\}}

\vspace{1em}
\noindent\textbf{Task}: From the list above, identify and summarize the \textbf{visually prominent and semantically significant anomalies} observed in the image.

You must analyze, consolidate, and explain each anomaly in a way that is \textbf{logical, detailed, and persuasive}, as if communicating to both experts and non-experts.

\vspace{1em}
\noindent\textbf{Instructions}:

\begin{enumerate}[leftmargin=2em]
    \item \textbf{Merge Similar or Redundant Anomalies}  
    \begin{itemize}
        \item Group phenomena sharing a common cause, concept, or visual effect.
        \item Avoid repetition by merging entries describing the same core issue.
    \end{itemize}

    \item \textbf{Resolve Contradictions Thoughtfully}
    \begin{itemize}
        \item If descriptions conflict, reconcile them using physical laws, biological plausibility, and visual logic.
        \item Summarize both viewpoints if both are partially valid.
    \end{itemize}

    \item \textbf{Filter Out Non-Visible or Insignificant Issues}
    \begin{itemize}
        \item Omit anomalies that are not visually apparent (e.g., minor texture noise).
        \item Focus on what is \textbf{clearly and prominently visible}.
    \end{itemize}

    \item \textbf{Justify with Real-World Logic}
    \begin{itemize}
        \item Support each anomaly with logical, physical, anatomical, or functional reasoning.
    \end{itemize}

    \item \textbf{Do Not Parrot the Input}
    \begin{itemize}
        \item Rephrase and reinterpret anomalies based on visual evidence and contextual understanding.
    \end{itemize}

    \item \textbf{Ensure Coverage}
    \begin{itemize}
        \item All input anomalies must be included, either directly or through consolidation.
    \end{itemize}
\end{enumerate}

\vspace{1em}
\noindent\textbf{Output Format}: Write a \textbf{numbered list}. For each entry, use the following structure:

\begin{Verbatim}[fontsize=\small]
@1. **Name**: [Descriptive title of the anomaly]  
- **Observed Phenomenon**:  
   - Describe what is visibly wrong in visual terms.  
   - Include positions, shapes, textures, or contextual oddities.  
   - Ensure clarity without needing to see the image.  
- **Reasoning**:  
   - Explain why this is implausible.  
   - Support with physical laws, anatomy, real-world logic, 
   or social context.  
- **Severity Score**: [0–100; 0 = fully unrealistic, 
   100 = fully realistic]
\end{Verbatim}

\vspace{0.5em}
\noindent\textbf{Example Output}:

\begin{enumerate}[leftmargin=2em]
    \item \textbf{Name}: Abnormal number of hands \\
    \textbf{Observed Phenomenon}: The individual on the left has \textit{two left hands}, one emerging from the elbow and overlapping with the sleeve. Both hands share identical orientation and lack anatomical continuity. \\
    \textbf{Reasoning}: Human anatomy allows one left and one right hand. Two left hands in such arrangement violate biological symmetry and visual plausibility. \\
    \textbf{Severity Score}: 5/100 (highly unrealistic)

    \item \textbf{Name}: Suspended chair without support \\
    \textbf{Observed Phenomenon}: A wooden chair is floating approximately 30 cm above the ground without visible support or shadows. \\
    \textbf{Reasoning}: Gravity requires contact or suspension; absence of legs, shadows, or wires defies physical realism. \\
    \textbf{Severity Score}: 10/100 (extremely unnatural)
\end{enumerate}

\vspace{1em}
\noindent\textbf{Final Notes}:
\begin{itemize}[leftmargin=2em]
    \item Output only the structured list in the format above.
    \item Think critically. Be precise, complete, and persuasive.
    \item Provide a human-understandable summary of core visual anomalies in the image.
\end{itemize}

\end{quote}

\noindent \textbf{Effectiveness:}
These final stages ensure that the output of the anomaly detection process is both precise and interpretable. The integration step reduces redundancy and focuses on the most significant anomalies, while the formatting step makes it easier to present and use the results for further analysis.

\noindent \textbf{Overall Design and Significance:}
The design of the anomaly detection agents is driven by the need for both local and global anomaly detection in AI-generated images. By breaking down the process into three main agents—Object Perceiver, Attribute Analyzer, and Relation Reasoner—each agent can focus on a specific aspect of the image, from identifying objects and their attributes to analyzing how they interact in a scene. This modular approach allows for greater flexibility and interpretability in the detection process. Furthermore, the integration and formatting stages ensure that the final output is structured and usable for real-world applications.

The multi-agent design enables a more nuanced and robust detection of semantic anomalies, making the framework capable of handling complex image generation models like those used in deepfake detection and other high-level image assessments.

\section{Comparative Analysis: Human vs. AnomAgent Annotations}
\label{appendix:agent_human_comparison}

To evaluate the effectiveness and consistency of our proposed anomaly detection pipeline, we conducted a comparative study between \textbf{AnomAgent} and \textbf{human annotators}. We used a benchmark set of \textbf{1,000 AI-generated images} with rich semantic content and potential visual inconsistencies.

\subsection{Annotation Task and Setup}

Each image was annotated independently by:
\begin{itemize}
    \item \textbf{Human annotators}, we select several annotators with advanced academic backgrounds in computer vision, specifically individuals holding master's or doctoral degrees..
    \item \textbf{AnomAgent}, using the pipeline described in Appendix~\ref{sec:agent_details}.
\end{itemize}

Annotators were instructed to identify any \textit{implausible, inconsistent, or unnatural visual features}, along with a brief description of the issue. AnomAgent, in contrast, automatically generated structured reports with descriptions, reasoning, and severity.

\begin{figure}[htbp]
    \centering
    \includegraphics[width=0.75\linewidth]{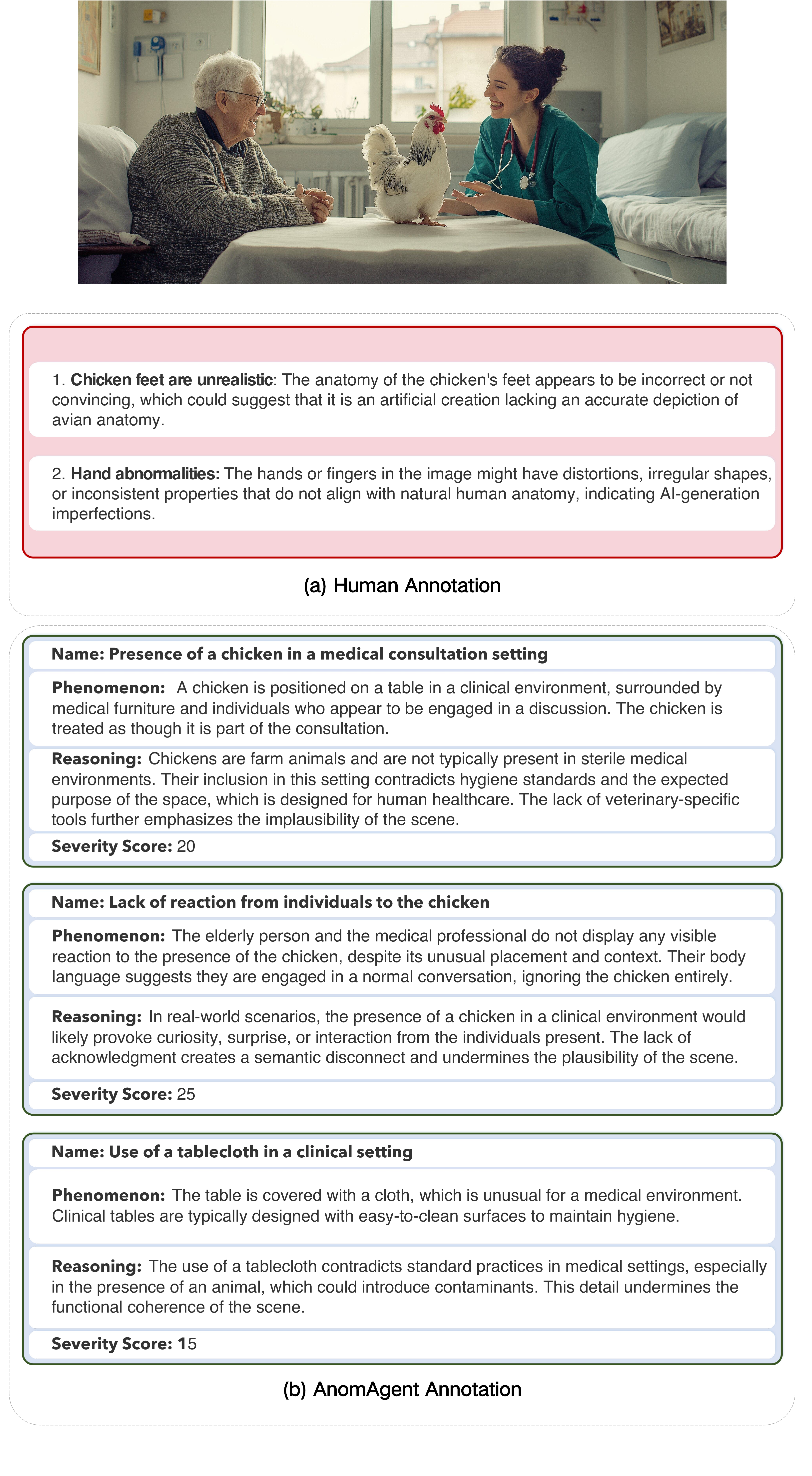}
    \caption{Comparison of human annotations and AnomAgent outputs. AnomAgent identifies more anomalies with detailed explanations.}
    \label{fig:agent_vs_human}
\end{figure}

\subsection{Qualitative Comparison}

Figure~\ref{fig:agent_vs_human} shows side-by-side comparisons of annotations from AnomAgent and human annotators. Each example highlights different strengths:
\begin{itemize}
    \item \textbf{Fine-grained detection}: AnomAgent captured subtle anomalies (e.g., missing shadows, unnatural finger joints) often missed by humans.
    \item \textbf{Consistency}: Across images with similar anomalies (e.g., floating objects), AnomAgent produced structurally similar descriptions, whereas human annotations varied in terminology and detail.
    \item \textbf{Reasoning support}: Human annotators often provided short descriptions without justification, while AnomAgent included structured, interpretable explanations.
\end{itemize}

For a dataset comprising 1000 images, human annotators identified 4884 anomalies, whereas AnomAgent detected 8290 anomalies. AnomAgent not only detects more anomalies, but does so with higher consistency and full interpretability with structured anomaly description.

\section{Comparison of BERTScore Backbones for Semantic Evaluation}
\label{appendix:bertscore_models}

Our semantic evaluation framework (Section~\ref{subsec:eval}) relies on BERTScore to compare structured textual anomaly explanations. However, BERTScore’s output depends heavily on the underlying language model used to compute embeddings. To assess the stability of our evaluation metrics, we benchmark several pre-trained language models as BERTScore backbones. This analysis tests whether the relative performance trends of vision-language models (VLMs) hold across different similarity functions.

In addition to the default \texttt{distilbert-base-uncased} model used in the main paper, we evaluate two widely used and diverse backbones:
\begin{itemize}
    \item \texttt{roberta-large-mnli}
    \item \texttt{google/mt5-large}
\end{itemize}

We use the exact same evaluation pipeline as in Section~\ref{subsec:eval}, computing \SemAP and \SemFOne metrics between human-verified ground truth anomalies and predictions from each model.

Table~\ref{tab:roberta-large-mnli} and Table~\ref{tab:google_mt5-large} present semantic detection and reasoning performance for all evaluated models across both backbones. 

The absolute values of \SemAP and \SemFOne vary across backbones. For example, \texttt{mt5-large} consistently yields lower scores due to its multilingual and less fine-grained English representations. In contrast, \texttt{roberta-large-mnli} produces more moderate values, balancing semantic sensitivity with surface-level alignment. However, the relative ranking of models—e.g., GPT-4o, AnomReasonor-7B, and Qwen2.5 variants—remains largely unchanged, confirming the robustness of our evaluation findings.

Metrics related to reasoning quality (\SemAPRea, \SemFOneRea) show greater variation across backbones than observation-only metrics. This aligns with the expectation that free-form reasoning texts introduce more lexical and structural variability, making similarity estimation more dependent on model semantics.

Across both backbones, \textit{AnomReasonor-7B} achieves top-tier performance, closely tracking or surpassing proprietary models on reasoning-related submetrics. This reinforces its strong generalization in structured explanation tasks and validates the effect of fine-tuning on our anomaly supervision.

While the choice of BERTScore model influences raw scores, the relative performance trends among models remain stable. Thus, our benchmark conclusions are not overly sensitive to the specific similarity function used—providing evidence of metric robustness and cross-backbone reliability.

\begin{table}[h]
\centering
\caption{Performance comparison using \texttt{roberta-large-mnli} as BERTScore backbone.}
\label{tab:roberta-large-mnli}
\resizebox{\textwidth}{!}{
\begin{tabular}{lccc|ccc}
\toprule
\textbf{Model} & \textbf{\SemAPObs} & \textbf{\SemAPRea} & \textbf{\SemAPFull} & \textbf{\SemFOneObs} & \textbf{\SemFOneRea} & \textbf{\SemFOneFull} \\
\midrule
LongVA-7B       & 0.2243 & 0.2002 & 0.2176 & 0.1372 & 0.1185 & 0.1325 \\
LLaVA-OV-7B     & 0.2487 & 0.2107 & 0.2358 & 0.1514 & 0.1269 & 0.1435 \\
Phi-3.5-Vision  & 0.2427 & 0.2129 & 0.2324 & 0.2074 & 0.1798 & 0.1984 \\
MiniCPM-V-2.6   & 0.2949 & 0.2621 & 0.2844 & 0.1417 & 0.1244 & 0.1366 \\
InternVL2-8B    & 0.2620 & 0.2297 & 0.2512 & 0.2674 & 0.2346 & 0.2563 \\
InternVL2.5-8B  & 0.2455 & 0.2079 & 0.2320 & 0.2179 & 0.1843 & 0.2062 \\
InternVL3-8B    & 0.3320 & 0.2811 & 0.3127 & 0.1405 & 0.1190 & 0.1332 \\
InternVL3-9B    & 0.2697 & 0.2313 & 0.2507 & 0.2738 & 0.2352 & 0.2542 \\
mPLUG-Owl3-7B   & 0.3102 & 0.2768 & 0.2984 & 0.0965 & 0.0857 & 0.0928 \\
Qwen2-VL-7B     & 0.3156 & 0.2749 & 0.3006 & 0.1032 & 0.0914 & 0.0989 \\
InternVL2-26B   & 0.2876 & 0.2532 & 0.2727 & 0.2760 & 0.2434 & 0.2621 \\
Qwen2.5-VL-7B   & 0.3190 & 0.2791 & 0.3017 & 0.2893 & 0.2531 & 0.2739 \\
Qwen2.5-VL-72B  & 0.3379 & 0.2906 & 0.3125 & 0.3033 & 0.2607 & 0.2809 \\
\midrule
Gemini-2.5-pro  & 0.2485 & 0.2290 & 0.2360 & 0.1330 & 0.1221 & 0.1264 \\
GPT-o3          & 0.2860 & 0.2578 & 0.2860 & 0.2834 & 0.2555 & 0.2834 \\
GPT-5           & 0.2322 & 0.2156 & 0.2207 & 0.2718 & 0.2526 & 0.2582 \\
GPT-4o          & 0.3434 & 0.2955 & 0.3083 & 0.3707 & 0.3188 & 0.3325 \\
\midrule
\rowcolor{gray!10}
AnomReasonor-7B & 0.3342 & 0.3197 & 0.3151 & 0.3243 & 0.3103 & 0.3059 \\

\bottomrule
\end{tabular}
}
\end{table}

\begin{table}[h]
\centering
\caption{Performance comparison using \texttt{google/mt5-large} as BERTScore backbone.}
\label{tab:google_mt5-large}
\resizebox{\textwidth}{!}{
\begin{tabular}{lccc|ccc}
\toprule
\textbf{Model} & \textbf{\SemAPObs} & \textbf{\SemAPRea} & \textbf{\SemAPFull} & \textbf{\SemFOneObs} & \textbf{\SemFOneRea} & \textbf{\SemFOneFull} \\
\midrule
LongVA-7B       & 0.0535 & 0.0635 & 0.0393 & 0.0315 & 0.0331 & 0.0214 \\
LLaVA-OV-7B     & 0.0993 & 0.1044 & 0.0834 & 0.0613 & 0.0599 & 0.0500 \\
Phi-3.5-Vision  & 0.1326 & 0.0853 & 0.0943 & 0.1092 & 0.0666 & 0.0743 \\
MiniCPM-V-2.6   & 0.1289 & 0.0926 & 0.0884 & 0.0598 & 0.0426 & 0.0407 \\
InternVL2-8B    & 0.1224 & 0.1243 & 0.1076 & 0.1252 & 0.1243 & 0.1089 \\
InternVL2.5-8B  & 0.1029 & 0.1024 & 0.0869 & 0.0911 & 0.0844 & 0.0743 \\
InternVL3-8B    & 0.1642 & 0.1764 & 0.1488 & 0.0661 & 0.0706 & 0.0596 \\
InternVL3-9B    & 0.1391 & 0.1121 & 0.1125 & 0.1428 & 0.1125 & 0.1143 \\
mPLUG-Owl3-7B   & 0.1173 & 0.1267 & 0.0964 & 0.0350 & 0.0379 & 0.0287 \\
Qwen2-VL-7B     & 0.1306 & 0.1393 & 0.1136 & 0.0403 & 0.0427 & 0.0348 \\
InternVL2-26B   & 0.1774 & 0.1417 & 0.1488 & 0.1701 & 0.1341 & 0.1420 \\
Qwen2.5-VL-7B   & 0.2034 & 0.1697 & 0.1775 & 0.1839 & 0.1531 & 0.1605 \\
Qwen2.5-VL-72B  & 0.2078 & 0.1798 & 0.1850 & 0.1864 & 0.1611 & 0.1662 \\
\midrule
Gemini-2.5-pro  & 0.1905 & 0.1402 & 0.1632 & 0.0989 & 0.0718 & 0.0839 \\
GPT-o3          & 0.1811 & 0.0991 & 0.1330 & 0.1793 & 0.0973 & 0.1311 \\
GPT-5           & 0.1724 & 0.0999 & 0.1346 & 0.2031 & 0.1177 & 0.1587 \\
GPT-4o          & 0.2708 & 0.2171 & 0.2387 & 0.2933 & 0.2352 & 0.2586 \\
\midrule
\rowcolor{gray!10}
AnomReasonor-7B & 0.2736 & 0.2612 & 0.2653 & 0.2655 & 0.2532 & 0.2574 \\

\bottomrule
\end{tabular}
}
\end{table}

\end{document}